\PassOptionsToPackage{table}{xcolor}
\documentclass{template}

\usepackage{mathtools}
\usepackage{multirow}
\usepackage{bm}
\usepackage{placeins}
\usepackage{etoolbox}

\usepackage{amsmath,amsfonts,bm}

\def\eqref#1{equation~\ref{#1}}

\def\1{\bm{1}}

\DeclareMathAlphabet{\mathsfit}{\encodingdefault}{\sfdefault}{m}{sl}
\SetMathAlphabet{\mathsfit}{bold}{\encodingdefault}{\sfdefault}{bx}{n}

\newcommand{\E}{\mathbb{E}}

\newcommand{\KL}{D_{\mathrm{KL}}}

\newcommand{\Cov}{\mathrm{Cov}}

\setboolean{logo}{true}
\definecolor{lineblue}{HTML}{0072B2}

\hypersetup{
  colorlinks=true,
  linkcolor=red,
  citecolor=Cerulean,
  filecolor=magenta,
  urlcolor=magenta,
  pdftitle={ResOPD: Tail Residualization for Sparse On-Policy Distillation},
  pdfauthor={Penghui Yang, Long Xing, Xuanlang Dai, Ziyu Liu, Kai Chen, Yuhang Zang},
  pdfkeywords={Large Language Models, On-Policy Distillation, Variance Reduction, Sparse Teacher Feedback}
}

\patchcmd{\maketitle}{\rule{\textwidth}{1pt}}{\rule{\linewidth}{1pt}}{}{
  \PackageError{resopd}{AILAB title rule patch failed}{Check template.cls.}
}

\newcommand{\method}{ResOPD}
\newcommand{\gzero}{HETS}
\newcommand{\ind}{\mathbb{I}}
\newcommand{\bgrad}{\bm{g}}
\newcommand{\bscore}{\bm{s}}
\newcommand{\btoken}{\bm{G}}
\newcommand{\bhead}{\bm{h}}
\newcommand{\bmean}{\bm{\mu}}
\newcommand{\bcontrol}{\bm{B}}
\newcommand{\bfluct}{\bm{C}}
\newcommand{\bresid}{\bm{R}}
\newcommand{\bz}{\bm{Z}}

\newcommand\blfootnote[1]{%
  \begingroup
  \renewcommand\thefootnote{}\footnote{#1}%
  \addtocounter{footnote}{-1}%
  \endgroup
}

\title{ResOPD: Tail Residualization for Sparse On-Policy Distillation}
\author[1,2]{Penghui Yang}
\author[3]{Long Xing}
\author[2,4]{Xuanlang Dai}
\author[1,2]{Ziyu Liu}
\author[2]{Kai Chen}
\author[2]{Yuhang Zang$^\dagger$}

\affil[1]{Shanghai Jiao Tong University}
\affil[2]{Shanghai Artificial Intelligence Laboratory}
\makeatletter
\renewcommand\AB@affilsep{,\protect\\\protect\Affilfont}
\makeatother
\affil[3]{University of Science and Technology of China}
\affil[4]{Fudan University}

\begin{abstract}
On-policy distillation (OPD) trains a student model on self-generated trajectories, but transmitting dense teacher distributions across long reasoning traces creates prohibitive communication and memory bottlenecks. Practical systems therefore rely on sparse teacher interfaces, typically transmitting either the sampled-token score or a small Top-$k$ distribution. However, this sparse setting faces a fundamental dilemma: sampled-token estimators are unbiased but suffer from severe gradient variance, whereas directly optimizing Top-$k$ objectives introduces systematic bias. To improve this trade-off, we propose ResOPD (\textbf{O}n-\textbf{P}olicy \textbf{D}istillation with Tail \textbf{Res}idualization), which provides unbiased full-vocabulary reverse KL gradient estimation under on-policy sampling, with substantial variance reduction under the same sparse payload in the evaluated settings. ResOPD aggregates the unobserved vocabulary into an observable coarse tail event, computes its exact aggregate gradient, and samples only the fine-grained within-tail residual, which requires no additional teacher queries or forward passes. Extensive experiments demonstrate that ResOPD substantially reduces gradient variance, stabilizes online training dynamics, and improves downstream performance across the evaluated settings. These results establish ResOPD as an efficient, plug-and-play variance reduction primitive for sparse on-policy distillation.
\end{abstract}

\begin{document}

\blfootnote{$^\dagger$ Corresponding author: Yuhang Zang (zangyuhang@pjlab.org.cn).}
\blfootnote{Code is available at \url{https://github.com/InternLM/ResOPD}.}

\maketitle

\section{Introduction}

On-policy distillation (OPD)~\citep{agarwal2024policy,gu2024minillm} transfers reasoning and generation ability from a strong teacher to a smaller student. Unlike offline distillation \citep{kim-rush-2016-sequence} or supervised fine-tuning, it trains the student on self-generated trajectories with token-level teacher supervision, which reduces exposure bias over long sequences. OPD is now a standard step in Large Language Model (LLM) post-training at frontier labs, including Thinking Machines \citep{lu2025onpolicydistillation}, GLM \citep{zeng2026glm}, and DeepSeek \citep{xu2026deepseek}.

\begin{figure}[t]
\centering
\includegraphics[width=1.0\textwidth]{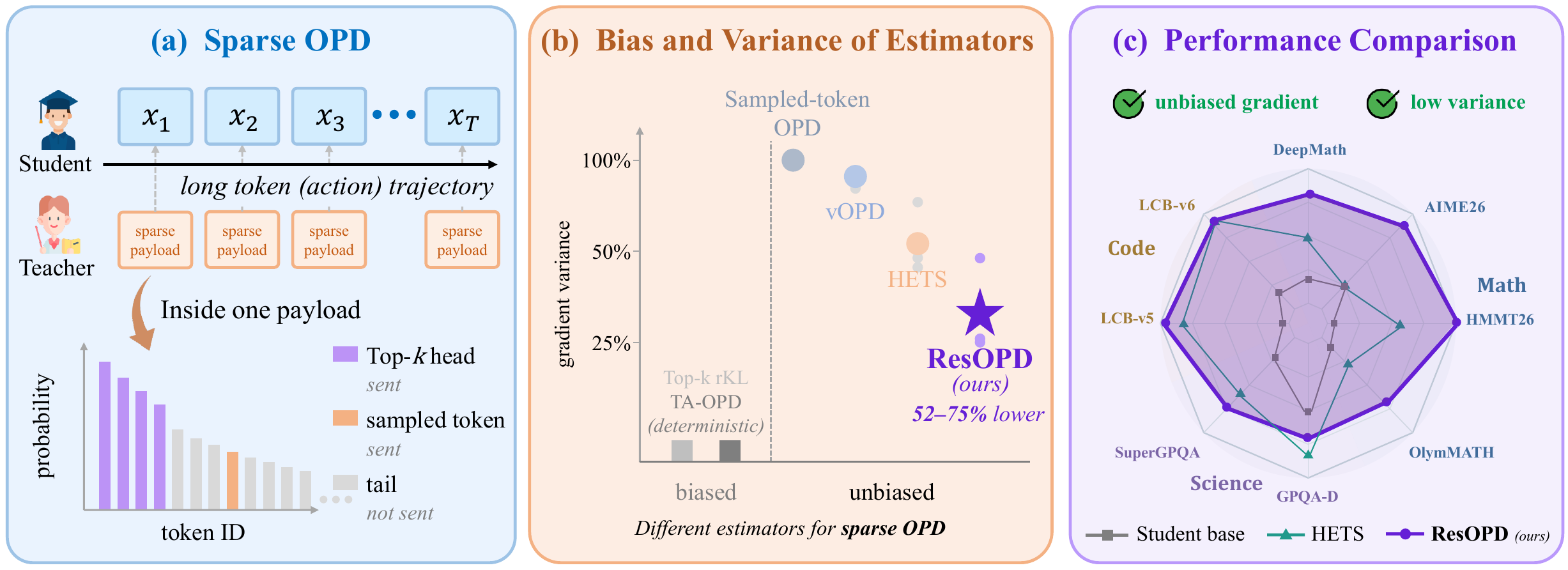}
\caption{(a) Sparse OPD: the teacher model only sends the Top-$k$ head probabilities and the sampled token score, leaving the tail distribution unobserved. (b) Bias vs.\ variance trade-off: while deterministic Top-$k$ surrogates suffer from systematic gradient bias, ResOPD provides an unbiased gradient estimator and substantially reduces gradient variance. (c) Downstream evaluation: the radar chart shows benchmark scores across mathematics, coding, and science; while the relative ranking varies by task, ResOPD achieves the best overall performance.}
\label{fig:estimator-anatomy}
\vspace{-4mm}
\end{figure}

Standard OPD \citep{agarwal2024policy} needs the teacher's full distribution at every generated token,
which is prohibitively costly for modern LLMs whose vocabularies can reach 250K tokens,
so the payload grows with vocabulary size per token and with rollout length per sequence. Practical training systems therefore expose a sparse teacher interface that, at each step, returns only the teacher's Top-$k$ tokens with their probabilities, along with the probability of the sampled token \citep{zhang2026ema,huang2026tail}. We refer to OPD trained through such an interface as \textbf{sparse OPD}. The difficulty is that the omitted tail is never observed: objectives defined directly on the Top-$k$ support, whether by truncation or renormalization \citep{zhu2026many} or by aggregating the omitted mass into a ghost token \citep{anshumann-etal-2025-sparse,huang2026tail}, are surrogates of the full-vocabulary reverse Kullback-Leibler (rKL) objective and carry a systematic gradient bias (Fig.~\ref{fig:estimator-anatomy}(b)).

Bias can be traded for variance: rather than shrinking the objective to the observed support, a second family of methods retains the full-vocabulary rKL objective and estimates its gradient from the sparse payload, remaining \textbf{unbiased but noisy}. However, existing estimators fail to fully exploit this payload. At one extreme, sampled-token OPD \citep{gu2024minillm,lu2025onpolicydistillation} queries only the rollout token's score, suffering from severe gradient variance due to single-sample Monte Carlo estimation (Fig.~\ref{fig:estimator-anatomy}(b)). At the other, methods with Top-$k$ access underutilize it: scalar-baseline variants such as vOPD \citep{oh2026kl} reduce the head to a detached baseline rather than evaluating its token-wise gradient analytically (Tab.~\ref{tab:mechanism}), while hybrid schemes such as \citet{zhang2026ema}, which we refer to as Head-Exact Tail-Sampled (HETS), evaluate the head exactly but treat the tail correction as entirely unknown despite knowing its aggregate mass. Consequently, each tail hit must compensate for the entire omitted mass, injecting large gradient spikes; even at 98.4\% head coverage, a 30K-token reasoning trace contains 460--470 such spikes, thousands per optimizer step (Sec.~\ref{sec:train_dynamics}).
Practitioners are thus left choosing between a biased objective and an unstable gradient, which causes sampled-token OPD and HETS to peak early and degrade later in training (Sec.~\ref{sec:math-opd}).

To resolve this trade-off, we analyze sparse OPD at the level of the gradient estimator. We first show that any estimator evaluating the head exactly, adding a prefix-dependent control vector weighted by the student's tail mass, and correcting on tail hits is unbiased for the full-vocabulary rKL gradient, with HETS as the special case of a zero control (Sec.~\ref{subsec:cfr}). Our covariance-trace decomposition reveals that variance splits into an invariant within-tail sampling term and a between-event term proportional to the control's distance from the true tail mean. This explains HETS's gradient spikes: its zero control pays for the full tail mean on every hit, whereas any control that better predicts this mean strictly reduces variance. Crucially, the residual masses outside the head define a coarse pseudo-token \citep{anshumann-etal-2025-sparse,huang2026tail} whose analytical gradient serves as a computable tail-mean predictor. Building on this, we propose \textit{\textbf{O}n-\textbf{P}olicy \textbf{D}istillation with Tail \textbf{Res}idualization} (\method{}): at every step it computes the exact gradient of the $(k+1)$-way coarse distribution, adding only a small residual correction on tail hits (Fig.~\ref{fig:framework}(a)). The resulting estimator is provably unbiased while retaining the same sparse teacher payload.

We evaluate \method{} in two settings. In exact frozen-prefix audits that enumerate every next-token action, \method{} lowers the gradient covariance trace by 52--75\% relative to sampled-token OPD under Top-$4$ feedback, below both HETS and a payload-matched vOPD baseline (Tab.~\ref{tab:mechanism}). In end-to-end distillation from Qwen3.5-27B to Qwen3.5-4B, it attains the best average score among the evaluated sparse OPD baselines on four mathematical reasoning benchmarks, 4.40 points above HETS under the same payload (Tab.~\ref{tab:math-terminal}),
and its average advantage over HETS consistently holds across every tested support size $k\in\{1,4,16,32\}$, with small margins (Tab.~\ref{tab:topk-breakdown}).

Our contributions are as follows:
(1) \textbf{Bias and variance analysis of sparse OPD.} We show that truncated, renormalized, and stopped-coefficient Top-$k$ rKL objectives are all biased surrogates of the full-vocabulary objective, and we derive a covariance-trace decomposition that separates within-tail sampling variance from between-event variance. This yields the exact condition under which a coarse tail control lowers variance relative to HETS, and a two-control family in which HETS, \method{}, and the concurrent SC-TA \citep{huang2026tail} appear as special cases.
(2) \textbf{\method{} estimator.} We propose an estimator that splits the full-vocabulary rKL gradient into an exact $(k+1)$-way coarse term, computed at every step from the Top-$k$ payload and the residual tail masses, and a within-tail residual correction applied only when the rollout token lands in the tail. The estimator is unbiased under on-policy sampling and requires no additional teacher queries or forward passes.
(3) \textbf{Empirical validation.} Using exact frozen-prefix gradient audits and end-to-end distillation settings, we show that \method{} reduces gradient covariance, lowers online residual moments during training, and improves downstream accuracy over the evaluated sparse OPD baselines.

\section{Related Work}

\paragraph{On-Policy Language-Model Distillation.}
On-policy distillation (OPD) mitigates exposure bias by training students on self-generated rollouts under dense teacher supervision \citep{gu2024minillm,agarwal2024policy}. Subsequent work explores skew divergences \citep{ko2024distillm}, adaptive targets \citep{jang2026stable,heo2026policy,jin2026entropy}, and diagnostic behaviors in complex reasoning regimes \citep{fu2026revisiting,zhu2026many}. However, full-vocabulary OPD incurs prohibitive communication and memory footprints during distributed rollout generation, severely hindering scalability.
\vspace{-4mm}
\paragraph{Sparse Divergences and Truncation Biases.}
To alleviate communication bottlenecks, distributed training frequently adopts sparse Teacher Top-$k$ payloads \citep{zhu2026many}. Nonetheless, naively renormalizing or truncating to the Top-$k$ support optimizes a biased surrogate that misallocates tail probability mass. While coarse-support approaches like TA-OPD \citep{huang2026tail} incorporate a ghost token \citep{anshumann-etal-2025-sparse} to aggregate omitted mass, they still optimize a surrogate $(k+1)$-way divergence that discards the teacher's fine-grained within-tail geometry.
\vspace{-4mm}
\paragraph{Stochastic Estimators and Variance Reduction.}
An alternative paradigm constructs stochastic estimators to recover the true full-vocabulary rKL gradient unbiasedly. Standard sampled-token estimators \citep{gu2024minillm,lu2025onpolicydistillation} suffer from extreme Monte Carlo variance, motivating variance-reduction baselines like vOPD \citep{oh2026kl}. HETS, a hybrid scheme proposed in EMA-PG \citep{zhang2026ema}, computes the Top-$k$ head analytically and samples only within the tail, yet still incurs severe gradient bursts upon tail hits, which destabilizes training dynamics.

\section{Methodology}
\label{sec:event}

\begin{figure}[t]
\centering
\includegraphics[width=0.8\textwidth]{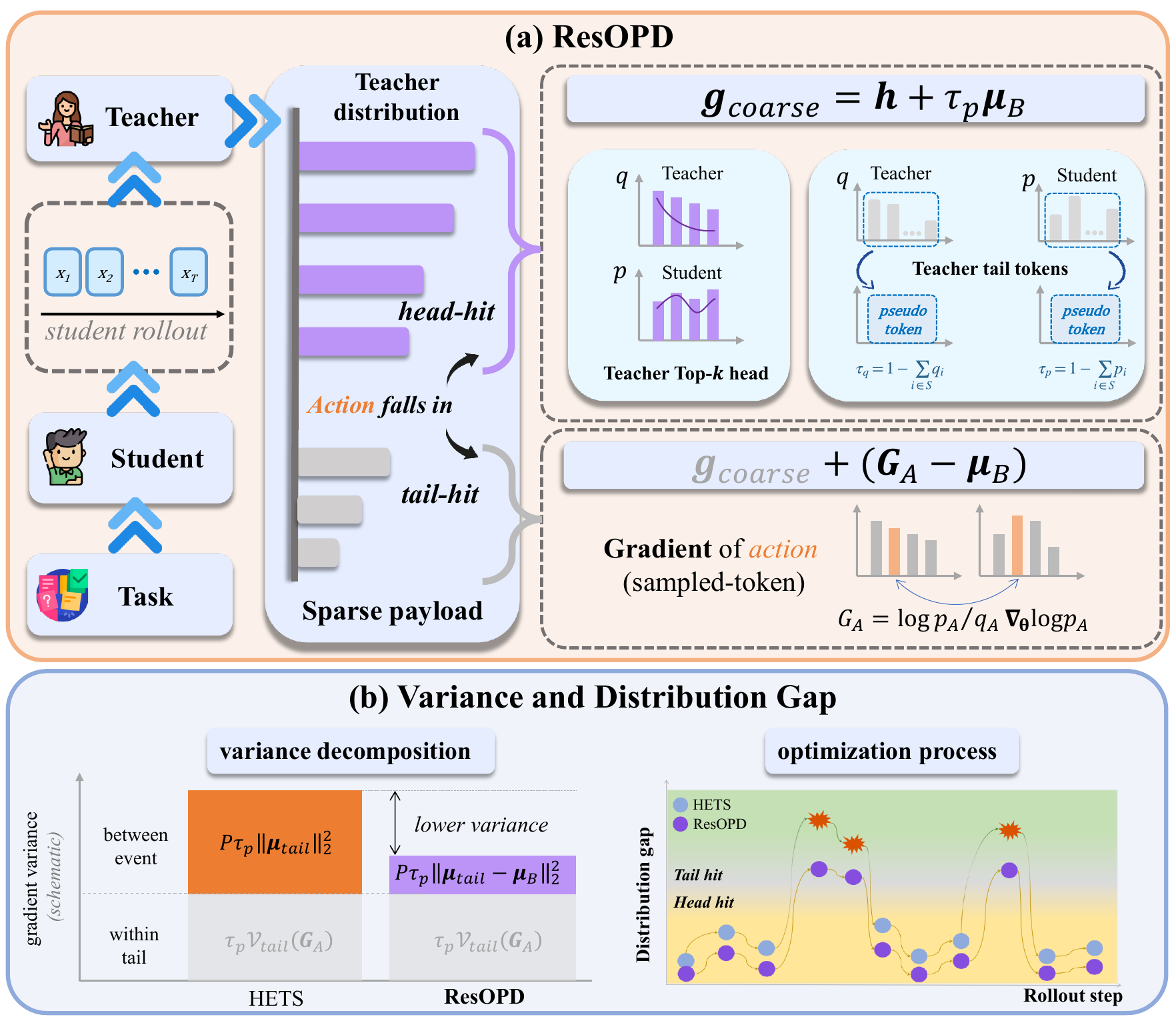}
\caption{
\textbf{Overview of ResOPD.}
(a) When the student sampled-token falls in the head, ResOPD uses only the exact coarse gradient; upon a tail hit, it adds a residual correction. (b) ResOPD suppresses between-event variance compared to HETS by centering tail samples around $\boldsymbol{\mu}_B$, achieving superior alignment with the teacher distribution across rollout steps.
}
\label{fig:framework}
\end{figure}

In this section, we present the theoretical foundations and algorithmic formulation of \textit{{O}n-{P}olicy {D}istillation with Tail {Res}idualization} (\method{}). 
We begin in Sec.~\ref{subsec:prelim} by establishing the full-vocabulary reverse-KL gradient and analyzing the trade-off of sparse teacher payloads, exposing how naive Top-$k$ heuristics induce objective bias and why hybrid estimators like HETS suffer from tail-sampling variance. 
In Sec.~\ref{subsec:cfr}, we formulate \method{} (illustrated in Fig.~\ref{fig:framework}), demonstrating how aggregate tail masses serve as an analytic control variate to achieve an unbiased, variance-reduced estimator. 
Sec.~\ref{subsec:evidence} describes the sampling assumptions and
connects theoretical covariance bounds to observable online diagnostics and downstream task performance.
Finally, Sec.~\ref{sec:unified} extends the analysis to a two-control family.



\subsection{Preliminaries: From sampled gradients to an exact sparse head}
\label{subsec:prelim}

At one fixed trajectory prefix, let $p(i)>0$ and $q(i)>0$ denote Student and Teacher
next-token probabilities over a shared vocabulary $\mathcal V$. Define
\begin{equation}
 \begin{aligned}
 r_i &\coloneqq \log\frac{p_i}{q_i}, &
 \bscore_i &\coloneqq \nabla_\theta\log p_i, &
 \btoken_i &\coloneqq r_i\bscore_i.
 \end{aligned}
 \label{eq:token-gradient}
\end{equation}
Because $\sum_i p_i\bscore_i=\bm 0$, the full-vocabulary reverse-KL gradient is
\begin{equation}
 \bgrad
 =\sum_i p_i\btoken_i
 =\E_{A\sim p}\!\left[\btoken_A\right].
 \label{eq:full-gradient}
\end{equation}
Here, $\btoken_A$ denotes the stopped-coefficient sampled-token OPD gradient. Equation~(\ref{eq:full-gradient}) demonstrates that this sampled-token gradient is an unbiased estimator of the full-vocabulary reverse-KL gradient.

\paragraph{Why direct Teacher Top-$k$ rKL is biased.}
Let $S$ be the fixed Teacher Top-$k$ support, with $0<k<|\mathcal V|$, $P=\sum_{i\in S}p_i$ and
$Q=\sum_{i\in S}q_i$. Two common scalar shortcuts have different biases:
\begin{equation}
 L_{\mathrm{renormalized}}
 =\KL\!\left(\frac{p_S}{P}\middle\|\frac{q_S}{Q}\right),
 \qquad
 L_{\mathrm{truncated}}=\sum_{i\in S}p_i\log\frac{p_i}{q_i}.
 \label{eq:topk-shortcuts}
\end{equation}
Renormalization depends only on the conditional distribution within $S$. And for every omitted logit $z_j$, $\partial L_{\mathrm{renormalized}}/\partial z_j=0$. Truncation retains the unnormalized head but introduces a boundary term:
\begin{equation}
 \frac{\partial L_{\mathrm{truncated}}}{\partial z_j}
 =-p_j\underbrace{\sum_{i\in S}p_i(r_i+1)}_{C_S},
 \qquad j\notin S.
 \label{eq:truncated-tail-push}
\end{equation}
When $C_S>0$, gradient descent increases every omitted logit.
We adopt the stopped-coefficient alternative \citep{zhu2026many}
$\bhead=\sum_{i\in S}p_i r_i\bscore_i$ in our experiments, which avoids that scalar boundary term but is
still biased by the omitted tail. Appendix~\ref{app:topk-bias} proves all three statements.

The interface returns full-normalized $q_i$ for $i\in S$ and $q_A$ for the
sampled action. Write
\begin{equation}
  \begin{gathered}
  \tau_p\coloneqq1-P,\qquad \tau_q\coloneqq1-Q,\\[-1pt]
  \bhead\coloneqq P\,\E[\btoken_A\mid A\in S] = \sum_{i\in S} p_i\btoken_i,
  \qquad
  \bmean_{\mathrm{tail}}\coloneqq
  \E\!\left[\btoken_A\mid A\notin S\right] = \frac{1}{\tau_p}\sum_{i\notin S} p_i\btoken_i.
  \end{gathered}
  \label{eq:head-tail-definitions}
\end{equation}
The head $\bhead$ is exactly known, while the fine-tail mean gradient is unavailable. With
$J=\ind\{A\notin S\}$, the prior Head-Exact Tail-Sampled estimator is
\begin{equation}
 \widehat{\bgrad}^{\mathrm{HETS}}
 =\bhead+J\btoken_A,
 \qquad
 \E\!\left[\widehat{\bgrad}^{\mathrm{HETS}}\right]
 =\bhead+\tau_p\bmean_{\mathrm{tail}}=\bgrad.
 \label{eq:g0}
\end{equation}
This \gzero{} baseline \citep{zhang2026ema} equals stopped Teacher Top-$k$ rKL
on head draws and adds the entire correction only on tail draws:
\begin{equation}
 \widehat{\bgrad}^{\mathrm{HETS}}
 =
 \begin{cases}
  \bhead, & A\in S,\\
  \bhead+\btoken_A, & A\notin S,
 \end{cases}
 \qquad
 \Pr(A\in S)=P.
 \label{eq:g0-rare-correction}
\end{equation}
Large $P$ therefore creates a rare-correction regime: most tokens use the
biased head-only update, while long minibatches still accumulate many sampled tail corrections.

\subsection{Tail residualization}
\label{subsec:cfr}

Top-$k$ also reveals aggregate tail masses.
Following the tail-bucketing approach in \citet{hubotter2026reinforcement} and \citet{huang2026tail}, collapsing the tail into a single pseudo-token yields
\begin{equation}
 \begin{aligned}
 b&\coloneqq\log\frac{\tau_p}{\tau_q}, &
 \bar{\bscore}_{\mathrm{tail}}
 &\coloneqq\E[\bscore_A\mid J=1]
   =\nabla_\theta\log\tau_p, &
 \bmean_B&\coloneqq b\,\bar{\bscore}_{\mathrm{tail}}.
 \end{aligned}
 \label{eq:bucket-vector}
\end{equation}
The exact gradient of the resulting complete coarse distribution is
\begin{equation}
 \bgrad_{\mathrm{coarse}}
 =\bhead+\tau_p\bmean_B.
 \label{eq:coarse-gradient}
\end{equation}
Here $\bmean_B$ predicts, but need not equal, the unavailable fine-tail mean. In fact, this formulation can be generalized to arbitrary baseline vectors: for $A\sim p$ and any prefix-dependent baseline vector $\bcontrol$,
\begin{equation}
\widehat{\bgrad}(\bcontrol)=\bhead+\tau_p\bcontrol+J(\btoken_A-\bcontrol),
\label{eq:residualization}
\end{equation}
is unbiased, i.e., $\E_{A\sim p}\!\left[\widehat{\bgrad}(\bcontrol)\right]=\bgrad$. Fig.~\ref{fig:framework} summarizes the construction.
HETS is the special case $\bcontrol=\bm0$. The expectation is explicit:
\begin{equation}
 \E[\widehat{\bgrad}(\bcontrol)]
 =\bhead+\tau_p\bcontrol
  +\tau_p(\bmean_{\mathrm{tail}}-\bcontrol)
 =\bgrad.
 \label{eq:residualization-expectation}
\end{equation}
The control may depend on the fixed prefix and current model probabilities,
but must be fixed with respect to the sampled action and stopped before
backward.
Furthermore, defining the total variance of any random vector $\bm{X}$ as $\mathcal{V}(\bm{X}) \coloneqq \operatorname{tr}(\operatorname{Cov}(\bm{X}))$, applying the law of total covariance yields the explicit variance of $\widehat{\bgrad}(\bm B)$:
\begin{equation}
\mathcal{V}(\widehat{\bgrad}(\bm B)) = \tau_p\mathcal{V}_{\mathrm{tail}}(\btoken_A)
+ P\tau_p\lVert\bmean_{\mathrm{tail}}-\bm B\rVert_2^2,
\label{eq:single-control-variance-1}
\end{equation}
where $\mathcal{V}_{\mathrm{tail}}(\btoken_A) \coloneqq \operatorname{tr}(\operatorname{Cov}(\btoken_A \mid A \in \mathrm{tail}))$.
Here, the first term $\tau_p\mathcal{V}_{\mathrm{tail}}(\btoken_A)$ captures the baseline-invariant within-tail sampling variance (weighted by the tail probability $\tau_p$), while the second term accounts for the between-event variance induced by the residual approximation error $\lVert\bmean_{\mathrm{tail}}-\bm B\rVert_2^2$, scaled by \(P\tau_p\), the variance of the binary head-versus-tail indicator.
To see this directly, condition on $J$: the head branch is deterministic,
the tail branch retains $\Cov_{\mathrm{tail}}(\btoken_A)$, and the two
conditional means differ by $\bmean_{\mathrm{tail}}-\bm B$. The law of total
covariance then gives Equation~(\ref{eq:single-control-variance-1}).

Equation~(\ref{eq:single-control-variance-1}) also shows that the variance of the estimator $\widehat{\bgrad}(\bcontrol)$ is minimized when the control vector is set to the true tail mean, i.e., $\bcontrol = \bmean_{\mathrm{tail}}$. However, since $\bmean_{\mathrm{tail}}$ is unavailable in practice, ResOPD substitutes it with the computable tail-bucket surrogate $\bmean_B$:
\begin{equation}
\widehat{\bgrad}^{\mathrm{ResOPD}}
=\bgrad_{\mathrm{coarse}}+J(\btoken_A-\bmean_B).
\label{eq:event}
\end{equation}

The predictable coarse correction now appears on every draw; only the
fine-tail residual remains rare.
ResOPD achieves strictly lower variance than HETS whenever $\lVert\bmean_{\mathrm{tail}}-\bmean_B\rVert_2 < \lVert\bmean_{\mathrm{tail}}\rVert_2$, i.e., whenever the tail-bucket vector $\bmean_B$ predicts the true tail mean better than zero, as observed in aggregate in our frozen-prefix audits. The complete derivation is given in Appendix~\ref{app:single-control}.

\subsection{Sampling assumptions and evidence boundaries}
\label{subsec:evidence}

The unbiasedness identities above assume $A\sim p_\theta$ at the fixed prefix.
Rollout filters change this sampling law. Our primary mathematics experiments (Sec.~\ref{sec:math-opd}) use
temperature $1.0$, rollout top-$p=0.95$, and rollout top-$k=20$, while the
full-support audits (Sec.~\ref{sec:exact-audit}) and cross-domain runs (Sec.~\ref{sec:cross-domain}) use the ancestral condition. Sec.~\ref{sec:ex-setup} specifies both, and
Sec.~\ref{sec:rollout-ablation} evaluates the effect of removing the filters.

The coarse-to-fine variance decomposition can also be characterized
under an arbitrary realized rollout distribution $\tilde p$. Let
$\tilde{\tau}=\Pr_{\tilde p}(A\notin S)$ and
$\tilde{\bm\mu}_{\mathrm{tail}}
=\mathbb{E}_{\tilde p}[\bm G_A\mid A\notin S]$.
For
$\hat{\bm g}(\bm B)=\bm h+\tau_p\bm B+
\mathbb{I}\{A\notin S\}(\bm G_A-\bm B)$,
the estimator still admits the exact decomposition
\begin{equation}
\mathcal V_{\tilde p}(\hat{\bm g}(\bm B))
=
\tilde{\tau}\mathcal V_{\tilde p,\mathrm{tail}}(\bm G_A)
+
(1-\tilde{\tau})\tilde{\tau}
\|\tilde{\bm\mu}_{\mathrm{tail}}-\bm B\|_2^2 .
\label{eq:realized-rollout-variance}
\end{equation}
Hence, for $0<\tilde{\tau}<1$, ResOPD has strictly lower sampling variance than HETS
whenever
$\|\tilde{\bm\mu}_{\mathrm{tail}}-\bm\mu_B\|_2
<\|\tilde{\bm\mu}_{\mathrm{tail}}\|_2$.
The expected update under this realized sampling law is
\begin{equation}
 \E_{\tilde p}[\widehat{\bgrad}(\bm B)]
 =\bhead+\tau_p\bm B
  +\tilde\tau(\tilde{\bm\mu}_{\mathrm{tail}}-\bm B).
 \label{eq:realized-rollout-mean}
\end{equation}
The variance identity remains exact under $\tilde p$, but it does not turn a
truncated rollout into an unbiased draw from $p_\theta$.
Crucially, truncation is confined strictly to sequence generation. All student quantities ($p_i$, $r_i$, $s_i$, $\tau_p$, and the coarse-tail terms) are faithfully recomputed via a fresh, untruncated forward pass during training.

Equation~(\ref{eq:single-control-variance-1}) characterizes estimator quality
at fixed prefixes. We test it by exact action enumeration
(Sec.~\ref{sec:exact-audit}). During online OPD, both policies and visited
prefixes change, while sparse Teacher feedback hides the fine tail.
We instead monitor residual second moments \(M_2\) as cheap mechanism-aligned online diagnostics. Their decline is consistent with reduced head-versus-tail event noise, but does not by itself estimate full gradient covariance.
Appendix~\ref{app:online-residual-moments} gives the exact definitions.

Nevertheless, variance reduction does not imply monotonic downstream accuracy gains. Lower covariance trace reduces stochastic optimization noise, but final performance is still shaped by training dynamics and decoding randomness, both of which are discretized by benchmark scoring. Thus, ResOPD can reliably suppress gradient noise without strictly outperforming HETS at every checkpoint. We therefore view theoretical covariance, online residual moments, and downstream accuracy as complementary signals across different stages of the learning pipeline.

\FloatBarrier
\subsection{A Unified View of Sparse Gradient Estimators}
\label{sec:unified}

ResOPD changes the between-event component of the HETS variance while
preserving its conditional within-tail randomness. This distinction provides
a useful way to compare sparse unbiased estimators: a second,
conditionally zero-mean control can act inside the tail. We present the
family, its oracle optimum, and the exact conditions governing the
ResOPD--SC-TA comparison here. The corresponding audits and downstream
results appear in Sec.~\ref{sec:control-comparison}.

\subsubsection{Separating between-event and within-tail controls}
\label{sec:joint-family}
\label{sec:event-scta}

The single-control family acts only on the head-versus-tail event component. A
second control can reduce the conditional variance that remains inside the
tail without changing the sparse Teacher payload. Define the conditionally
zero-mean score fluctuation
$\bfluct_A=b(\bscore_A-\bar{\bscore}_{\mathrm{tail}})$, so that
$\E[\bfluct_A\mid J=1]=\bm0$. For any finite prefix-measurable $\bm B$ and
scalar $\gamma$, both stopped before drawing $A$, define

\begin{equation}
 \widehat{\bgrad}(\bm B,\gamma)
 =\bhead+\tau_p\bm B
 +J\!\left(\btoken_A-\bm B-\gamma\bfluct_A\right).
 \label{eq:unified-main}
\end{equation}
Equation~(\ref{eq:unified-main}) satisfies
\begin{equation}
 \begin{aligned}
 \E\!\left[\widehat{\bgrad}(\bm B,\gamma)\right]
 &=\bhead+\tau_p\bm B
   +\tau_p(\bmean_{\mathrm{tail}}-\bm B)
   -\gamma\tau_p\E[\bfluct_A\mid J=1]\\
 &=\bhead+\tau_p\bmean_{\mathrm{tail}}
  =\bgrad.
 \end{aligned}
 \label{eq:family-unbiased-proof}
\end{equation}
Conditioned on the head event its value is $\bhead+\tau_p\bm B$, while conditioned
on the tail event its mean is
$\bhead+\tau_p\bm B+\bmean_{\mathrm{tail}}-\bm B$. Their difference is
$\bmean_{\mathrm{tail}}-\bm B$, while only the tail branch retains conditional randomness.
The law of total covariance therefore gives
\begin{equation}
 \mathcal V\!\left(\widehat{\bgrad}(\bm B,\gamma)\right)
 =\tau_p\mathcal V_{\mathrm{tail}}(\btoken_A-\gamma\bfluct_A)
 +P\tau_p\lVert\bmean_{\mathrm{tail}}-\bm B\rVert_2^2.
 \label{eq:two-level-variance}
\end{equation}
This factorization proves that the two controls are orthogonal in trace
covariance: $\bm B$ changes only the between-event term and $\gamma$ only
the within-tail term. We call Equation~(\ref{eq:unified-main}) the
\emph{Unified} family: the first coordinate generalizes the single control in
Equation~(\ref{eq:residualization}), while the second subtracts a
known-zero-mean within-tail fluctuation. If
$\mathcal V_{\mathrm{tail}}(\bfluct_A)>0$, their joint
oracle is
\begin{equation}
 \bm B^\star=\bmean_{\mathrm{tail}},\qquad
 \gamma^\star=
 \frac{\operatorname{tr}\!\left(
       \Cov_{\mathrm{tail}}(\btoken_A,\bfluct_A)\right)}
      {\mathcal V_{\mathrm{tail}}(\bfluct_A)},
 \label{eq:joint-oracle}
\end{equation}
and completing the square in both coordinates yields
\begin{equation}
 \begin{aligned}
 &\mathcal V\!\left(\widehat{\bgrad}(\bm B,\gamma)\right)
 -\mathcal V\!\left(
   \widehat{\bgrad}(\bm B^\star,\gamma^\star)\right)\\
 &\qquad
 =P\tau_p\lVert\bm B-\bmean_{\mathrm{tail}}\rVert_2^2
 +\tau_p\mathcal V_{\mathrm{tail}}(\bfluct_A)
  (\gamma-\gamma^\star)^2.
 \end{aligned}
 \label{eq:unified-regret-main}
\end{equation}
If $\mathcal V_{\mathrm{tail}}(\bfluct_A)=0$, $\gamma$ is immaterial and the first term
remains exact. Importantly, neither oracle coordinate is available online:
$\bm B^\star$ requires the full Teacher tail, while the per-prefix $\gamma^\star$
requires its conditional covariance. ResOPD instead makes the
simple but effective choice $(\bmean_B,0)$, and the global fitted coefficient is a
separate offline diagnostic.

\begin{table}[t]
\caption{Coordinates in the two-control family.}
\label{tab:estimator-map}
\centering
\begin{tabular*}{0.7\linewidth}{@{\extracolsep{\fill}} lccc @{}}
\toprule
Unbiased estimator & $\bm B$ & $\gamma$ & Role \\
\midrule
HETS & $\bm 0$ & $0$ & structural baseline \\
ResOPD & $\bmean_B$ & $0$ & our estimator \\
SC-TA & $\bmean_B$ & $1$ & boundary estimator \\
Joint oracle & $\bmean_{\mathrm{tail}}$ & $\gamma^\star$ & unavailable optimum \\
\bottomrule
\end{tabular*}
\end{table}
\subsubsection{ResOPD and SC-TA as endpoints}
\label{sec:gamma}

We now specialize the two-level family to $\bm B=\bmean_B$ and connect it to sample-corrected TA-OPD (SC-TA) from the concurrent work of \citet{huang2026tail}. Their primary method, TA-OPD, augments the Teacher Top-$k$ support with a tail pseudo-token and optimizes the resulting coarse $(k+1)$-way reverse-KL objective. SC-TA is subsequently derived as a sample-corrected variant that restores full-vocabulary unbiasedness. We complement this construction with a gradient-level characterization within our two-control family. In particular, we derive the exact pointwise relation between ResOPD and SC-TA, characterize their covariance ordering as a function of the within-tail control coefficient $\gamma$, identify the trace-optimal $\gamma^\star$, and evaluate these predictions across natural prompt distributions.

\paragraph{Exact fine-tail decomposition.}
On a tail draw,
\begin{equation}
 r_A
 =\log\frac{p_A}{q_A}
 =\underbrace{\log\frac{\tau_p}{\tau_q}}_{b}
 +\log\frac{p_A/\tau_p}{q_A/\tau_q}.
\end{equation}
Using $\bmean_B=b\bar{\bscore}_{\mathrm{tail}}$ therefore gives
\begin{equation}
 \btoken_A
 =\bmean_B
 +\underbrace{b(\bscore_A-\bar{\bscore}_{\mathrm{tail}})}_{\bfluct_A}
 +\underbrace{(r_A-b)\bscore_A}_{\bresid_A},
 \qquad
 \E[\bfluct_A\mid J=1]=\bm 0.
 \label{eq:tail-exact-decomposition}
\end{equation}
The terms have different roles. $\bmean_B$ is the coarse tail-bucket prediction; $\bfluct_A$ is a known-zero-mean fluctuation around the conditional Student
score; and $\bresid_A$ records the unresolved Student--Teacher conditional
tail-shape mismatch.

With $\bgrad_{\mathrm{coarse}}=\bhead+\tau_p\bmean_B$, the specialization of
Equation~(\ref{eq:unified-main}) is
\begin{equation}
 \widehat{\bgrad}_\gamma
 =\bgrad_{\mathrm{coarse}}
 +J\!\left[\bresid_A+(1-\gamma)\bfluct_A\right].
 \label{eq:event-scta-family}
\end{equation}
Equation~(\ref{eq:family-unbiased-proof}) already proves that every fixed
$\gamma$ targets $\bgrad$. Its two named endpoints are
\begin{equation}
 \begin{aligned}
 \widehat{\bgrad}^{\mathrm{ResOPD}}
 &=\widehat{\bgrad}_{\gamma=0}
  =\bgrad_{\mathrm{coarse}}+J(\bresid_A+\bfluct_A),\\
 \widehat{\bgrad}^{\mathrm{SC}}
 &=\widehat{\bgrad}_{\gamma=1}
  =\bgrad_{\mathrm{coarse}}+J\bresid_A.
 \end{aligned}
\end{equation}
Hence their signed pointwise relationship is
\begin{equation}
 \boxed{
 \widehat{\bgrad}^{\mathrm{SC}}
 =\widehat{\bgrad}^{\mathrm{ResOPD}}-J\bfluct_A
 }
 \label{eq:event-scta-difference}
\end{equation}
The difference has zero expectation but is generally nonzero on tail actions.

\paragraph{Variance optimum and endpoint certificate.}
Let $\bz_0 = \bresid_A + \bfluct_A = \btoken_A - \bmean_B$. Decomposing the total variance into the common between-event component and the $\gamma$-dependent within-tail component yields
\begin{equation}
  \begin{aligned}
    \mathcal{V}(\widehat{\bgrad}_\gamma)
    &= \tau_p \mathcal{V}_{\mathrm{tail}}(\bz_0 - \gamma \bfluct_A) + P\tau_p \lVert\bmean_{\mathrm{tail}} - \bmean_B\rVert_2^2, \\
    \mathcal{V}_{\mathrm{tail}}(\bz_0 - \gamma \bfluct_A)
    &= \mathcal{V}_{\mathrm{tail}}(\bz_0) - 2\gamma \operatorname{tr}\!\left(\Cov_{\mathrm{tail}}(\bz_0, \bfluct_A)\right) + \gamma^2 \mathcal{V}_{\mathrm{tail}}(\bfluct_A).
  \end{aligned}
  \label{eq:event-scta-total-var}
\end{equation}
When $\mathcal{V}_{\mathrm{tail}}(\bfluct_A) > 0$, completing the square with respect to $\gamma$ identifies the trace optimum:
\begin{equation}
  \begin{aligned}
    \gamma^\star
    &= \frac{\operatorname{tr}\!\left(\Cov_{\mathrm{tail}}(\bz_0, \bfluct_A)\right)}{\mathcal{V}_{\mathrm{tail}}(\bfluct_A)}, \\
    \mathcal{V}(\widehat{\bgrad}_\gamma) - \mathcal{V}(\widehat{\bgrad}_{\gamma^\star})
    &= \tau_p \mathcal{V}_{\mathrm{tail}}(\bfluct_A) (\gamma - \gamma^\star)^2.
  \end{aligned}
  \label{eq:event-scta-regret}
\end{equation}
Evaluating the quadratic regret at the boundary points $\gamma = 1$ (SC-TA) and $\gamma = 0$ (ResOPD) reveals that SC-TA achieves strictly lower trace covariance than ResOPD if and only if $\gamma^\star > 1/2$. Substituting $\bz_0 = \bresid_A + \bfluct_A$ expands the optimal coefficient as
\begin{equation}
  \gamma^\star = 1 + \frac{\operatorname{tr}\!\left(\Cov_{\mathrm{tail}}(\bresid_A, \bfluct_A)\right)}{\mathcal{V}_{\mathrm{tail}}(\bfluct_A)}.
  \label{eq:event-scta-gamma-residual}
\end{equation}
Consequently, SC-TA outperforms ResOPD exactly when $\operatorname{tr}\!\left(\Cov_{\mathrm{tail}}(\bresid_A, \bfluct_A)\right) > -\mathcal{V}_{\mathrm{tail}}(\bfluct_A)/2$. By the Cauchy--Schwarz inequality, a sufficient condition is
\begin{equation}
  \mathcal{V}_{\mathrm{tail}}(\bresid_A) < \mathcal{V}_{\mathrm{tail}}(\bfluct_A)/4.
  \label{eq:event-improvement}
\end{equation}
In particular, when the conditional tail shapes match, the residual vanishes ($\bresid_A = \bm{0}$), yielding $\gamma^\star = 1$. These conditions serve as distribution-dependent certificates rather than guaranteeing a universal variance ordering.

\FloatBarrier
\section{Experiments}
\label{sec:experiments}

We evaluate three complementary questions: whether residualization reduces
exact conditional gradient covariance, whether its observable effects
persist during online training, and whether the resulting policies improve
downstream performance. We first specify the protocols, then present
frozen-prefix audits, mathematics results and diagnostics, cross-domain
transfer, and controlled ablations of support size, rollout sampling,
learning rate, and the within-tail control.

\subsection{Experimental Setup}
\label{sec:ex-setup}

\paragraph{Offline mechanism audit.}
We freeze Qwen3.5 \citep{qwen3.5} models and sample one 64-token trajectory per prompt at selected frozen Math
prefixes, measuring covariance trace and Noise-to-Signal Ratio:
NSR=$\frac{\mathcal{V}(\hat{\bgrad})}{\|\bgrad_{\text{full-rKL}}\|_2^2}$.
Experiments cover five settings: GSM8K \citep{cobbe2021gsm8k} (32 prompts) across 0.8B/4B, 4B/27B and 9B/27B pairs; all 33 HMMT26 prompts \citep{dekoninck2026matharena} (4B/27B); and 64 prompts sampled from DeepMath-103K \citep{deepmath} (4B/27B).

\paragraph{Online OPD training for mathematical reasoning.}
We distill the thinking-enabled Qwen3.5-4B student from the Qwen3.5-27B teacher. Starting from DeepMath-103K, we filter for challenging competition problems after deduplication and benchmark contamination removal. To ensure an appropriate teacher-student capability gap, we retain prompts where the teacher succeeds twice while the student fails at least once across sampled rollouts. This yields 1,276 unique problems, partitioned into 1,024 for training and 252 for conditional held-out evaluation. Training runs for 3 epochs (192 steps), saving intermediate checkpoints at steps 6, 32, 64, 128, and 192.
\vspace{-4mm}

\paragraph{Evaluation.}
Our evaluation spans four benchmarks: the DeepMath held-out set, AIME26 (30 problems) \citep{aime26}, HMMT26 (33 problems) \citep{dekoninck2026matharena}, and the English split of OlymMATH (200 problems) \citep{sun2025challengingboundariesreasoningolympiadlevel}. Given the limited problem counts in AIME26 and HMMT26, we report the average over 16 sampled responses per problem (avg@16) to mitigate evaluation variance, while DeepMath held-out and OlymMATH use avg@2. In addition to terminal performance, we evaluate all intermediate saved checkpoints to compute the Area Under the Curve (AUC), quantifying learning dynamics and training efficiency. We compute normalized trapezoidal AUC over the saved checkpoint
trajectory, and average the benchmark-wise scores and AUCs for the aggregate
columns. Full configurations for both training and evaluation are collected in
Appendix~\ref{app:hyperparameters}.
\vspace{-4mm}

\paragraph{Realized rollout distributions.}
Following the Qwen3.5 official recipe, the primary mathematics rollouts and evaluations
use temperature $1.0$, top-$p=0.95$, and top-$k=20$.
Truncating low-probability tokens helps avoid degenerate prefixes that can
destabilize OPD and Teacher feedback~\citep{fu2026revisiting}.
The Student probabilities used in training are recomputed by a fresh,
untruncated forward pass. As explained in Sec.~\ref{subsec:evidence}, this
setup retains a well-defined sampling-variance comparison but does not
satisfy the idealized full-rKL unbiasedness assumption.
We therefore also report a matched full-support mathematics ablation
(Sec.~\ref{sec:rollout-ablation}) and cross-domain training with
top-$p=1.0$, top-$k=-1$ (Sec.~\ref{sec:cross-domain}).


\subsection{Exact frozen-prefix covariance audits}
\label{sec:exact-audit}

\begin{table}[t]
\caption{
Exact frozen-prefix covariance trace and NSR. Covariance traces of student logit-gradients are computed via full action enumeration and averaged across rollouts; NSR normalizes each trace by the squared $L_2$ norm of the full-rKL gradient. Sparse estimators ($\text{vOPD}_{\text{T-Top-}k}$, HETS, ResOPD) share an identical teacher budget of $k=4$,
with $\text{vOPD}_{\text{T-Top-}k}$ adopting a detached conditional-KL baseline over this support.
Best results per row are in \textbf{bold}, and second-best are \underline{underlined}.
}
\label{tab:mechanism}
\centering
\small
\setlength{\tabcolsep}{2.8pt}
\renewcommand{\arraystretch}{1.18}

\begin{tabular*}{\linewidth}{@{\extracolsep{\fill}}ll cccc cccc@{}}
\toprule
\multirow{2}{*}{\textbf{Dataset}}
& \multirow{2}{*}{\textbf{Models}}
& \multicolumn{4}{c}{\textbf{Covariance Trace} $\downarrow$}
& \multicolumn{4}{c}{\textbf{NSR} $\downarrow$} \\
\cmidrule(lr){3-6} \cmidrule(l){7-10}
&
& \shortstack{Sampled} & \shortstack{$\text{vOPD}_{\text{T-Top-}k}$} & \shortstack{HETS} & \shortstack{ResOPD}
& \shortstack{Sampled} & \shortstack{$\text{vOPD}_{\text{T-Top-}k}$} & \shortstack{HETS} & \shortstack{ResOPD} \\
\midrule
GSM8K    & 0.8B\,/\,4B  & 2.146 & 1.730 & \underline{1.559} & \textbf{1.018} & 5.422 & 4.370 & \underline{3.940} & \textbf{2.573} \\
GSM8K    & 9B\,/\,27B   & 0.215 & 0.192 & \underline{0.107} & \textbf{0.068} & 6.202 & 5.560 & \underline{3.101} & \textbf{1.955} \\
GSM8K    & 4B\,/\,27B   & 0.333 & 0.297 & \underline{0.183} & \textbf{0.108} & 6.058 & 5.400 & \underline{3.335} & \textbf{1.964} \\
HMMT26   & 4B\,/\,27B   & 0.208 & 0.189 & \underline{0.092} & \textbf{0.052} & 6.589 & 5.977 & \underline{2.914} & \textbf{1.630} \\
DeepMath & 4B\,/\,27B   & 0.183 & 0.169 & \underline{0.087} & \textbf{0.047} & 7.352 & 6.785 & \underline{3.500} & \textbf{1.907} \\
\bottomrule
\end{tabular*}
\end{table}

Tab.~\ref{tab:mechanism} reports the exact covariance trace and NSR across five frozen-prefix panels. By analytically enumerating every next-token action at each Student-visited prefix, we isolate the conditional sampling variance of each estimator without decoding noise or optimizer feedback. Under the panel-level aggregation used in the table, both metrics exhibit a consistent ordering across every panel: $\text{ResOPD} < \text{HETS} < \text{vOPD} < \text{sampled-token OPD}$ (lower is better). Relative to sampled-token OPD, the aggregate covariance trace is reduced by $7.7\%\text{--}19.4\%$ for payload-matched Teacher-Top-$k$ vOPD, $27.3\%\text{--}55.8\%$ for \gzero{}, and $52.5\%\text{--}75.0\%$ for \method{}.

While Teacher-Top-$k$ vOPD serves as a competitive baseline in our end-to-end experiments (Tab.~\ref{tab:math-terminal}), its detached scalar control fails to exploit the structural head--tail decomposition, leaving the entire action vector sampled at every step. Consequently, its variance reduction remains modest compared to HETS and ResOPD. Because HETS and ResOPD share the identical payload budget, exact Top-4 head integration, and conditional target structure, their contrast directly isolates the gain from coarse-tail residualization without baseline confounds. We therefore focus our primary ablations on the HETS-to-ResOPD progression.

As the support size $k$ increases, the exact Top-$k$ head absorbs more probability mass, causing the residual tail opportunity $P\tau_p$ to contract. Accordingly, the marginal covariance reduction of \method{} over \gzero{} gradually narrows. The full support-size sweep in Appendix~\ref{app:fixed-prefix-sweep} tests this communication--variance trade-off directly.

\subsection{End-to-end mathematical reasoning}
\label{sec:math-opd}

\begin{table}[t]
\caption{
Qwen3.5-27B-to-4B OPD results. DeepMath held-out and OlymMATH are evaluated using avg@2, while AIME26 and HMMT26 adopt avg@16 due to their smaller problem sets.
}
\label{tab:math-terminal}
\centering
\small
\setlength{\tabcolsep}{3.5pt}
\renewcommand{\arraystretch}{1.18}

\newcolumntype{Y}{>{\centering\arraybackslash}X}

\begin{tabularx}{\linewidth}{l YYYYY Y}
\toprule
\multirow{2}{*}[-2ex]{\textbf{Method}}
& \multicolumn{4}{c}{\textbf{Benchmark Accuracy (\%)}}
& \multicolumn{2}{c}{\textbf{Aggregate (\%)}} \\
\cmidrule(lr){2-5} \cmidrule(l){6-7}
& \shortstack{DeepMath\\{\scriptsize held-out (avg@2)}}
& \shortstack{AIME26\\{\scriptsize(avg@16)}}
& \shortstack{HMMT26\\{\scriptsize(avg@16)}}
& \shortstack{OlymMATH\\{\scriptsize(avg@2)}}
& \shortstack{Avg.\\Score}
& \shortstack{Avg.\\AUC} \\
\midrule

\rowcolor{gray!18}\multicolumn{7}{@{}l}{\textit{\textbf{$\blacktriangledown$ Original Models}}} \\
Qwen3.5-4B Student   & 48.41 & 56.25 & 35.23 & 31.50 & 42.85 & -- \\
Qwen3.5-27B Teacher  & 89.29 & 80.62 & 62.88 & 63.25 & 74.01 & -- \\

\midrule
\rowcolor{gray!18}\multicolumn{7}{@{}l}{\textit{\textbf{$\blacktriangledown$ Alternative Sparse OPD (Biased)}}} \\
Teacher Top-$k$ fKL  & 50.20 & 54.58 & 35.04 & 34.50 & 43.58 & 42.80 \\
Teacher Top-$k$ rKL  & 51.79 & 55.83 & 34.66 & 36.75 & 44.76 & 44.63 \\
TA-OPD               & 58.13 & 58.96 & \underline{39.58} & 36.00 & 48.17 & \underline{47.94} \\

\midrule
\rowcolor{gray!18}\multicolumn{7}{@{}l}{\textit{\textbf{$\blacktriangledown$ Unbiased rKL Estimators}}} \\
Sampled-token OPD    & 48.21 & 57.50 & 36.17 & 34.25 & 44.03 & 46.69 \\
vOPD$_{\text{T-Top-}k}$ & \underline{58.93} & \textbf{60.00} & 39.20 & \underline{37.25} & \underline{48.85} & 47.49 \\
\gzero{}             & 54.76 & 56.25 & 37.88 & 34.00 & 45.72 & 46.18 \\

\rowcolor{lineblue!15}
\textbf{\method{}} (\textit{ours}) & \textbf{60.91} & \underline{59.38} & \textbf{39.96} & \textbf{40.25} & \textbf{50.12} & \textbf{47.96} \\
\bottomrule
\end{tabularx}
\end{table}

Tab.~\ref{tab:math-terminal} reports the primary distillation results:
\method{} achieves the strongest aggregate performance among all evaluated
OPD baselines, attaining both the highest terminal average score (50.12\%)
and learning-trajectory AUC (47.96\%).
This favorable training trajectory also extends to external mathematical
competitions. Across AIME26, HMMT26, and OlymMATH, competing baselines exhibit
pronounced post-peak degradation during later stages of training, whereas
\method{} avoids severe late-stage degradation and recovers to its best
aggregate performance at the terminal checkpoint. Step-wise evaluations over
intermediate checkpoints are provided and analyzed in Appendix~\ref{app:math-learning-trajectories}.

\subsubsection{Training dynamics}
\label{sec:train_dynamics}

\begin{table}[t]
\centering
\small
\caption{\textbf{Aggregated online diagnostics across training} (lower is better). Values represent the average over all 192 optimization steps. $\Delta$ denotes the relative percentage reduction of \method{} compared to HETS.}
\label{tab:stable-dynamics}
\renewcommand{\arraystretch}{1.18}
\setlength{\tabcolsep}{6pt}
\begin{tabular}{lccccc}
\toprule
\textbf{Diagnostic Metric} & \textbf{Top-$k$ rKL} & \textbf{$\text{vOPD}_{\text{T-Top-}k}$} & \textbf{HETS} & \textbf{ResOPD (\textit{ours})} & \textbf{$\Delta$ (\%)$\downarrow$} \\
\midrule
Support-mass gap ($\times10^{-3}$)          & 11.17 & 3.53  & 5.17  & \textbf{3.50}  & $-32.3\%$ \\
$|$Tail log-mass mismatch$|$                 & 0.715 & 0.858 & 0.723 & \textbf{0.322} & $-55.5\%$ \\
Token residual $M_2$ ($\times10^{-3}$)      & --    & \textbf{10.55} & 11.73 & {10.67} & $-9.0\%$  \\
Trajectory residual $M_2$ ($\times10^{-5}$) & --    & 2.76  & 3.50  & \textbf{2.65}  & $-24.3\%$ \\
\bottomrule
\end{tabular}
\vspace{-3mm}
\end{table}

\begin{figure}[t]
\centering
\includegraphics[width=0.85\textwidth]{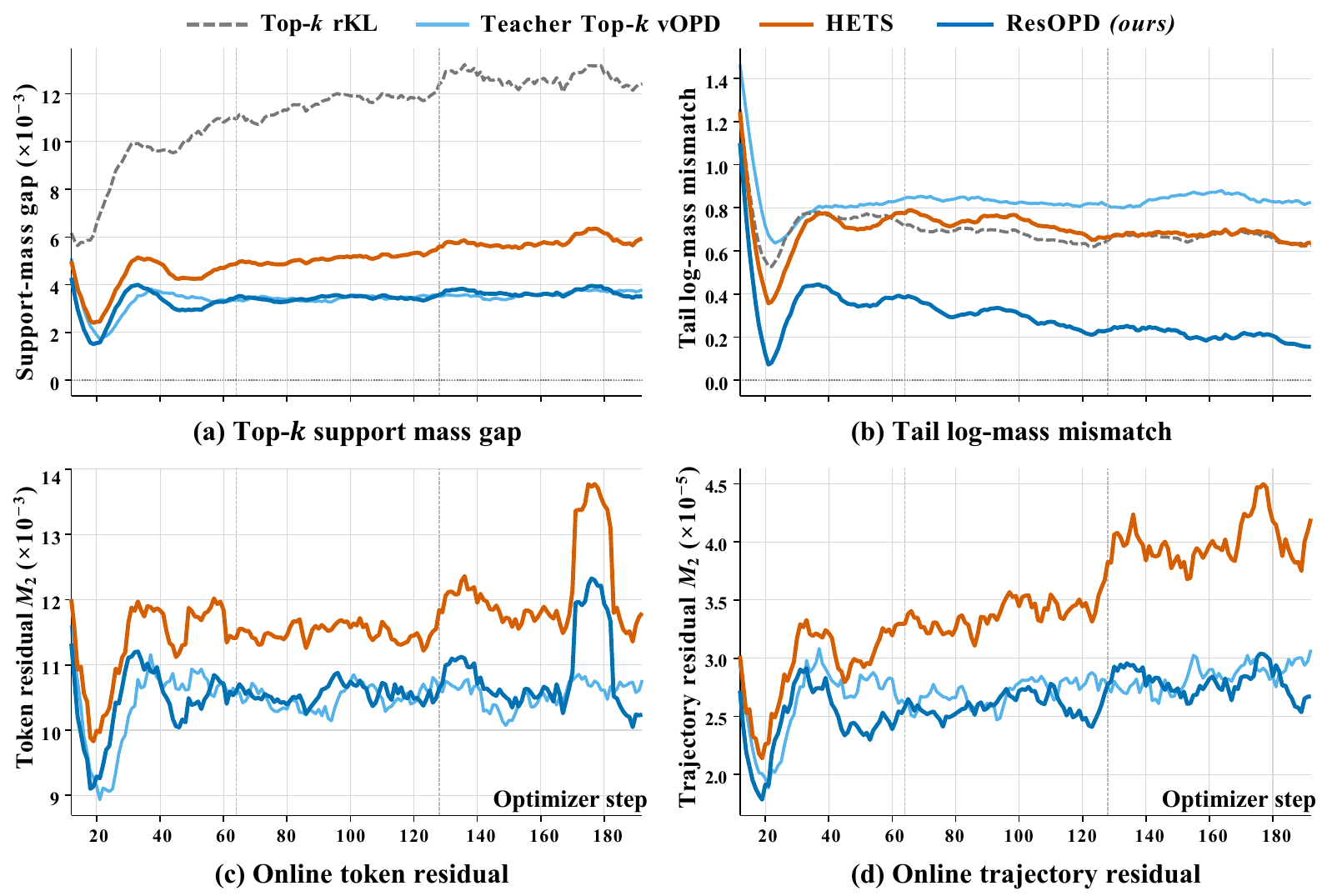}
\caption{\textbf{Training-time diagnostics} (64 steps/epoch; lower is better).
\textbf{Panels (a) and (b)} illustrate the student--teacher mismatch in Top-$k$ support mass and tail log-mass across vOPD, HETS, ResOPD, and direct Teacher Top-$k$ rKL.
\textbf{Panels (c) and (d)} compare vOPD, HETS and ResOPD via the squared residual between the binary Top-$k$ indicator and its student-predicted probability, evaluated at the token level and trajectory level, respectively. Detailed formulations are provided in Appendix~\ref{app:online-residual-moments}.}
\label{fig:training-dynamics-event-g0}
\vspace{-6mm}
\end{figure}

As illustrated in Tab.~\ref{tab:stable-dynamics} and Fig.~\ref{fig:training-dynamics-event-g0}, \method{} consistently suppresses both student--teacher mass discrepancies and stochastic residual fluctuations relative to HETS on average over 192 training steps. Across the matched HETS--\method{} runs, \method{} reduces the support-mass gap by $32.3\%$ and the tail log-mass mismatch by $55.5\%$.
Although Teacher-Top-$k$ vOPD exhibits a comparable support-event $M_2$, this scalar residual on self-generated trajectories omits centered log-ratio and score geometry, and exact gradient covariance comparisons on shared frozen prefixes (Tab.~\ref{tab:mechanism}) confirm ResOPD's structural advantage.

This provides a mechanistic explanation for the optimization behavior of \method{}. Under HETS, the full fine-tail gradient correction is concentrated on rare tail-hit events. Despite approximately $98.4\%$ head coverage, a $30\text{K}$-token reasoning trace still contains 460--470 such corrections, accumulating to thousands per optimizer step. By subtracting the payload-available tail-bucket component $\bmean_B$, \method{} redistributes the predictable between-event component while preserving the full-rKL expectation under the on-policy condition in Sec.~\ref{subsec:evidence}. Downstream accuracy remains noisier because it depends on discrete answer correctness, but the completion-only Teacher NLL audit below shows consistently higher teacher likelihood of student-generated trajectories for \method{}. Together with the matched endpoint and curve-AUC results, these diagnostics connect the estimator-level covariance reduction to a smoother optimization path and improved downstream performance.

\vspace{-4mm}
\subsubsection{Behavioral fidelity}
\label{sec:math-eval-diagnostics}

Beyond task-specific accuracy metrics (e.g., avg@2 and avg@16 on downstream math benchmarks), monitoring the student's negative log-likelihood under the teacher model provides an intrinsic, task-agnostic readout of distillation fidelity. Specifically, a lower Teacher NLL indicates that the student samples receive higher likelihood under the teacher.

\begin{figure}[htbp]
\centering
\includegraphics[width=0.9\textwidth]{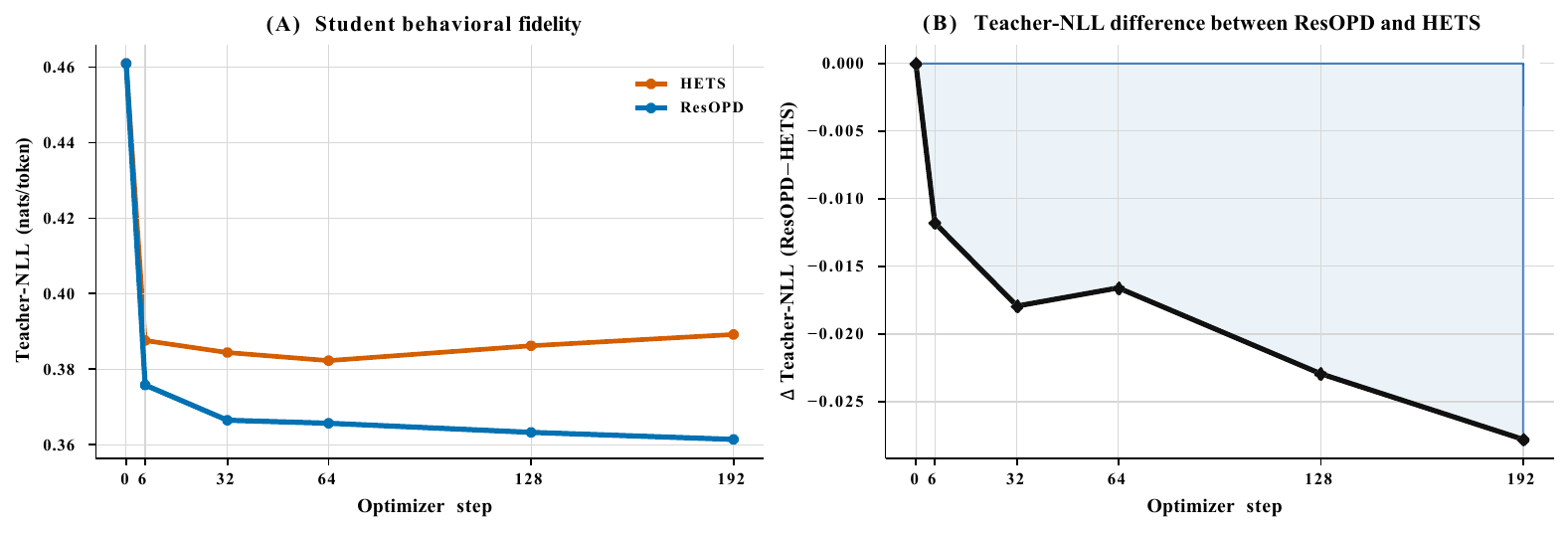}
\vspace{-2mm}
\caption{\textbf{Teacher negative log-likelihood (NLL) across training in the Math OPD setup.}
Evaluations are computed over the four core benchmark families with identical weights to the main suite: AIME26, HMMT26, DeepMath held-out, and OlymMATH. Lower values indicate higher fidelity to the Teacher distribution.}
\label{fig:teacher-nll}
\end{figure}

Fig.~\ref{fig:teacher-nll} tracks this metric across the matched Math run. As illustrated in Panel~A, while both estimators begin around $0.46$\,nats/token, ResOPD achieves a steeper drop by step 6 and sustains a monotonic decline down to $0.36$\,nats/token by step 192. In contrast, HETS plateaus around $0.385$\,nats/token and displays slight drift in later stages. The difference $\Delta\text{Teacher-NLL (ResOPD} - \text{HETS)}$ in Panel~B remains strictly negative across all optimizer steps, widening to over $-0.027$\,nats/token. This demonstrates that ResOPD consistently preserves higher behavioral fidelity to the teacher than HETS, offering evidence of higher teacher-likeness, complementary to accuracy.


\subsection{Cross-domain transfer to science and coding}
\label{sec:cross-domain}

\begin{table}[h]
\caption{
Distillation results from Qwen3-30B-A3B-Instruct to Qwen3-4B. Evaluations use avg@10 for GPQA-D and avg@3 for SuperGPQA, LCB-v5, and LCB-v6.
AUC integrates performance across the fixed training trajectory (Base/10/25/50/100).
}
\label{tab:qwen3-transfer}
\centering
\small
\setlength{\tabcolsep}{3.5pt}
\renewcommand{\arraystretch}{1.18}

\newcolumntype{Y}{>{\centering\arraybackslash}X}

\begin{tabularx}{\linewidth}{l YYYY YY}
\toprule
\multirow{2}{*}[-2ex]{\textbf{Method}}
& \multicolumn{2}{c}{\textbf{Science Benchmarks (\%)}}
& \multicolumn{2}{c}{\textbf{Code Benchmarks (\%)}}
& \multicolumn{2}{c}{\textbf{Aggregate (\%)}} \\
\cmidrule(lr){2-3} \cmidrule(lr){4-5} \cmidrule(l){6-7}
& \shortstack{GPQA-D\\{\scriptsize(avg@10)}}
& \shortstack{SuperGPQA\\{\scriptsize(avg@3)}}
& \shortstack{LCB-v5\\{\scriptsize(avg@3)}}
& \shortstack{LCB-v6\\{\scriptsize(avg@3)}}
& \shortstack{Avg.\\Score}
& \shortstack{Avg.\\AUC} \\
\midrule

\rowcolor{gray!18}\multicolumn{7}{@{}l}{\textit{\textbf{$\blacktriangledown$ Original Models}}} \\
Student             & 45.25          & 32.46          & 21.76          & 24.38          & 30.96          & -- \\
Teacher             & 56.06          & 48.87          & 45.51          & 41.71          & 48.04          & -- \\

\midrule
\rowcolor{gray!18}\multicolumn{7}{@{}l}{\textit{\textbf{$\blacktriangledown$ Unbiased rKL Estimators}}} \\
Sampled-token OPD   & \textbf{50.30} & \underline{36.87} & 43.31          & 39.43          & 42.48          & 41.21 \\
HETS                & \underline{48.89} & 36.10          & \underline{46.31} & \textbf{43.24} & \underline{43.63} & \underline{42.13} \\

\rowcolor{lineblue!15}
\textbf{ResOPD} (\textit{ours}) & 47.12          & \textbf{37.54} & \textbf{48.90} & \underline{43.05} & \textbf{44.15} & \textbf{42.65} \\
\bottomrule
\end{tabularx}
\end{table}

To further evaluate the generalization and cross-domain effectiveness of ResOPD, we follow \citet{heo2026policy} and extend our investigation to science and coding tasks. Our experimental protocol largely mirrors that of \citet{heo2026policy}: we distill from a Qwen3-30B-A3B-Instruct \citep{yang2025qwen3} teacher into a Qwen3-4B student, training on a 30k-sample mixture compiled from OpenMathReasoning \citep{moshkov2025aimo}, OpenScienceReasoning-2 \citep{nvidia2025opensciencereasoning2}, and OpenCodeReasoning \citep{ahmad2025opencodereasoning}. The run uses full-support ancestral rollouts with temperature $1.0$, top-$p=1.0$, and top-$k=-1$, without thinking mode. For evaluation, we benchmark the code domain on LiveCodeBench \citep[LCB;][]{jain2025livecodebench} v5 (167 problems) and v6 (175 problems). For the science domain, we evaluate on GPQA-Diamond \citep[GPQA-D, 198 problems;][]{rein2023gpqa} and SuperGPQA \citep{du2026supergpqa}, adopting the exact 650-problem subset used in \citet{heo2026policy}.

Tab.~\ref{tab:qwen3-transfer} reports the evaluation results. Overall, ResOPD achieves the strongest aggregate results, leading in both terminal average score ($44.15\%$) and trajectory AUC ($42.65\%$). Although ResOPD and HETS nominally surpass the teacher on code tasks (e.g., ResOPD's $48.90\%$ vs.\ $45.51\%$ on LCB-v5), paired bootstrap 95\% CIs cross zero (e.g., ResOPD$-$Teacher spans $[-2.48, +5.33]$ on LCB-v6), indicating on-par performance rather than statistically significant super-teacher generalization.
Across training scales, ResOPD maintains consistent optimization advantages (e.g., reducing trajectory residual $M_2$ by $9.5\%$--$12.6\%$ and gradient norms by $14.6\%$--$23.7\%$) and exhibits directional AUC gains over HETS on code ($+0.48$ and $+0.73$ points), even though Sampled-token OPD yields the strongest performance on science tasks (detailed training checkpoint curves are provided in Appendix~\ref{app:cross-qwen3}).
\vspace{-2mm}

\subsection{Online support-size ablation}
\label{sec:online-support}

\begin{table}[t]
\caption{Performance under different Top-$k$ support sizes (matched one-epoch budget). Evaluation setups follow Tab.~\ref{tab:math-terminal} (avg@16 for AIME26/HMMT26 and avg@2 for OlymMATH). $\Delta$ denotes the average improvement of ResOPD over HETS.}
\label{tab:topk-breakdown}
\centering
\small
\renewcommand{\arraystretch}{1.1}

\begin{tabular*}{0.80\textwidth}{@{\extracolsep{\fill}}cc cccc c@{}}
\toprule
\textbf{$k$} & \textbf{Method} & \textbf{AIME26} & \textbf{HMMT26} & \textbf{OlymMATH} & \textbf{Average} & \textbf{$\Delta$} \\
\midrule
\multirow{2}{*}{1}
& HETS & \textbf{60.62} & 37.69          & 34.75          & 44.35          & \multirow{2}{*}{$+0.11$} \\
& ResOPD  & 56.88          & \textbf{39.02} & \textbf{37.50} & \textbf{44.47} & \\
\addlinespace[3pt]
\multirow{2}{*}{4}
& HETS & 59.17          & 36.93          & 34.50          & 43.53          & \multirow{2}{*}{$+1.01$} \\
& ResOPD  & \textbf{60.00} & \textbf{37.88} & \textbf{35.75} & \textbf{44.54} & \\
\addlinespace[3pt]
\multirow{2}{*}{16}
& HETS & 57.92          & \textbf{39.39} & 35.75          & 44.35          & \multirow{2}{*}{$+0.70$} \\
& ResOPD  & \textbf{59.38} & 39.02          & \textbf{36.75} & \textbf{45.05} & \\
\addlinespace[3pt]
\multirow{2}{*}{32}
& HETS & 59.79          & \textbf{39.02} & 36.75          & 45.19          & \multirow{2}{*}{$+1.56$} \\
& ResOPD  & \textbf{61.67} & 38.07          & \textbf{40.50} & \textbf{46.75} & \\
\bottomrule
\end{tabular*}
\end{table}

\begin{figure}[t]
\centering
\includegraphics[width=0.9\textwidth]{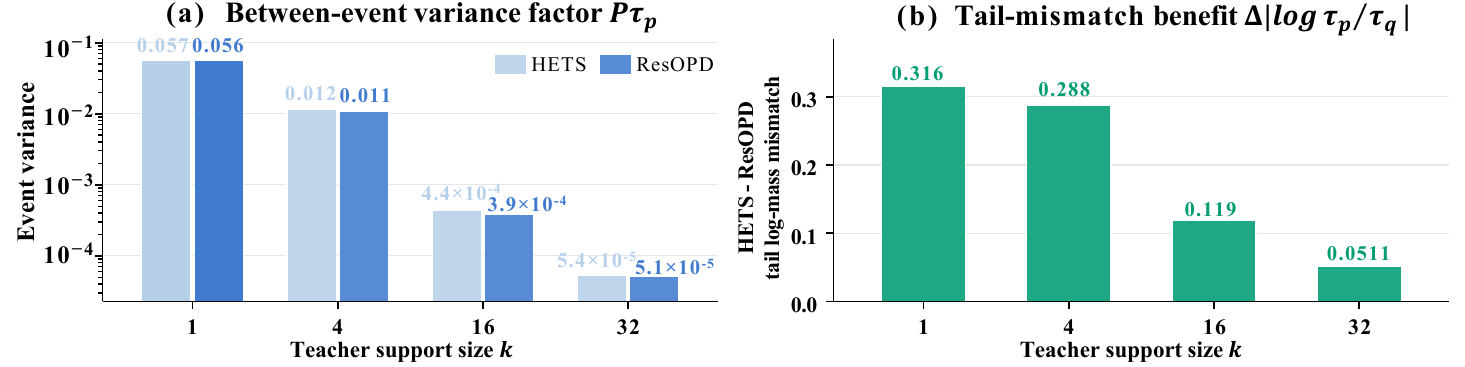}
\caption{Online diagnostic metrics across teacher Top-$k$ support sizes.
\textbf{(a)} Between-event variance factor $P\tau_p$ for HETS and ResOPD, characterizing the scaling coefficient of the head-versus-tail partition.
\textbf{(b)} Tail log-mass mismatch reduction ($\text{HETS} - \text{ResOPD}$, measured via $\Delta|\log(\tau_p / \tau_q)|$).}
\label{fig:topk-ablation}
\vspace{-4mm}
\end{figure}

Appendix~\ref{app:fixed-prefix-sweep} examines the impact of the support size $k$ with frozen prefixes, reporting the exact covariance trace and NSR. In contrast, here we provide an end-to-end online ablation of OPD. Specifically, Tab.~\ref{tab:topk-breakdown} and Fig.~\ref{fig:topk-ablation} show the results across teacher support sizes $k \in \{1, 4, 16, 32\}$.
As shown in Tab.~\ref{tab:topk-breakdown}, ResOPD consistently outperforms HETS across all tested support sizes in average downstream accuracy, with improvements ranging from $+0.11$ percentage points at $k=1$ up to $+1.56$ points at $k=32$.
However, while downstream task scores consistently favor ResOPD, the intrinsic variance reduction benefit offered by ResOPD is expected to diminish as $k$ increases, since a larger teacher support naturally captures more probability mass and renders tail corrections less frequent.

Fig.~\ref{fig:topk-ablation} validates this mechanism on the runs from Tab.~\ref{tab:topk-breakdown}.
In Fig.~\ref{fig:topk-ablation}(a), HETS and ResOPD exhibit nearly identical variance factors $P\tau_p$,  both dropping by over three orders of magnitude from $k=1$ to $32$. This directly scales the theoretical
ResOPD--HETS variance contrast alongside tail-mean approximation quality.
Meanwhile, Fig.~\ref{fig:topk-ablation}(b) shows that ResOPD consistently lowers tail log-mass mismatch across all supports,
with the largest margin in tight-support regimes.

\subsection{Rollout sampling: default filters versus full support}
\label{sec:rollout-ablation}
While the idealized ancestral sampling condition $A\sim p_\theta$ is strictly enforced in both our exact enumeration audits (Sec.~\ref{sec:exact-audit}) and the cross-domain transfer experiments (Sec.~\ref{sec:cross-domain}), we also verify for completeness that the primary math results do not depend on rollout truncation. Specifically, we evaluate the first training epoch under full-support sampling ($\text{top-}p=1.0$, $\text{top-}k=-1$) while keeping all other training hyperparameters identical.
As shown in Tab.~\ref{tab:full-support-first-epoch}, removing the rollout filters lowers AUC for all three evaluated methods. Nevertheless, ResOPD consistently maintains its advantage over HETS and the naive baseline across both benchmark accuracy and trajectory AUC. In addition, we observe that removing rollout filters yields longer reasoning trajectories and a higher frequency of hitting the response token limit during training.


\subsection{Learning-rate sensitivity}
\label{sec:lr-sensitivity}

\begin{table}[t]
\centering
\small
\caption{\textbf{First-epoch Math performance under default vs. full-support rollouts.}
Benchmark scores and Avg. denote step-64 accuracy (\%); $\text{AUC}$ (\%) measures the normalized learning curve area over steps $\{0, 6, 32, 64\}$.
\textit{Default}: top-$p=0.95$, top-$k=20$; \textit{Full}: top-$p=1.0$, top-$k=-1$.}
\vspace{-2mm}
\label{tab:full-support-first-epoch}
\setlength{\tabcolsep}{5pt}
\renewcommand{\arraystretch}{1.15}
\begin{tabular}{llcccccc}
\toprule
\textbf{Rollout} & \textbf{Method} & \textbf{DeepMath} & \textbf{AIME26} & \textbf{HMMT26} & \textbf{OlymMATH} & \textbf{Avg.} & \textbf{AUC} \\
\midrule
\multirow{3}{*}{Default}
& Naive Top-$k$ rKL & 53.17 & 56.04 & 35.42 & 35.00 & 44.91 & 45.45 \\
& HETS              & 53.17 & 59.17 & 36.93 & 34.50 & 45.94 & 46.12 \\
& ResOPD            & \textbf{54.76} & \textbf{60.00} & \textbf{37.88} & \textbf{35.75} & \textbf{47.10} & \textbf{47.57} \\
\midrule
\multirow{3}{*}{Full}
& Naive Top-$k$ rKL & 45.44 & 55.42 & 35.80 & 34.25 & 42.72 & 44.96 \\
& HETS              & 49.80 & 57.50 & \textbf{37.88} & 35.50 & 45.17 & 45.68 \\
& ResOPD            & \textbf{57.14} & \textbf{60.42} & \textbf{37.88} & \textbf{36.00} & \textbf{47.86} & \textbf{46.79} \\
\bottomrule
\end{tabular}
\end{table}

\paragraph{Experimental setup and evaluation protocol.}
To evaluate optimization robustness under aggressive step sizes, we conduct a controlled learning-rate sweep ($k=4$) identical to the primary Math experimental setup, except that training is compressed to 1~epoch (64 optimizer steps). All runs share a batch size of 16, a 32K response budget, and identical sampling configurations with the primary experiments. Checkpoints at steps 0, 6, 32, and 64 are evaluated under the Macro-4 protocol (averaging 16 responses per problem for AIME26 and HMMT26, and 2 responses for DeepMath held-out and OlymMATH).
\vspace{-4mm}

\begin{table}[t]
\centering
\small
\caption{\textbf{Learning-rate sensitivity of HETS vs.\ ResOPD.} $\Delta$ denotes the improvement of ResOPD over HETS.}
\vspace{-2mm}
\label{tab:lr-sensitivity}
\renewcommand{\arraystretch}{1.18}
\setlength{\tabcolsep}{7pt}
\begin{tabular}{@{}l cc c cc c@{}}
\toprule
\multirow{2}{*}{\textbf{Learning Rate}}
& \multicolumn{3}{c}{\textbf{Final Score (\%)}}
& \multicolumn{3}{c}{\textbf{Trajectory AUC (\%)}} \\
\cmidrule(lr){2-4} \cmidrule(l){5-7}
& HETS & ResOPD & $\Delta$
& HETS & ResOPD & $\Delta$ \\
\midrule
$1\times10^{-6}$ & 45.94 & \textbf{47.10} & $+1.16$ & 46.12 & \textbf{47.57} & $+1.45$ \\
$2\times10^{-6}$ & 45.94 & \textbf{46.61} & $+0.67$ & 45.69 & \textbf{46.17} & $+0.48$ \\
$5\times10^{-6}$ & 42.54 & \textbf{46.35} & \textbf{+3.81} & 43.88 & \textbf{47.77} & \textbf{+3.89} \\
\bottomrule
\end{tabular}
\vspace{-4mm}
\end{table}

\paragraph{Performance resilience across learning rates.}
As shown in Tab.~\ref{tab:lr-sensitivity}, ResOPD demonstrates substantially stronger resilience at elevated learning rates. When scaling the learning rate to $5\times10^{-6}$, HETS suffers severe degradation, with its final score dropping from $45.94\%$ to $42.54\%$ ($-3.40\%$) and its trajectory AUC declining by $2.24\%$. In contrast, ResOPD retains consistent performance ($46.35\%$ final score, $47.77\%$ AUC), widening its lead over HETS to $+3.81\%$ in final score and $+3.89\%$ in AUC.
Examining the sub-benchmark breakdown at this aggressive rate, ResOPD's advantage is consistently positive across all four constituent benchmarks at step 64: $+3.96\%$ on AIME26, $+2.08\%$ on HMMT26, $+6.94\%$ on DeepMath held-out, and $+2.25\%$ on OlymMATH. We provide detailed training dynamics across different learning rates in Appendix~\ref{app:lr-sensitivity}.
\vspace{-4mm}

\paragraph{Stability boundaries and failure modes.}
When the learning rate is scaled further to $1\times10^{-5}$, both methods diverge rapidly, triggering heavy gradient clipping ($4/6$ steps for HETS and $5/6$ steps for ResOPD) and severe response-length collapse within six optimizer steps. While ResOPD does not shift the ultimate divergence boundary ($1\times10^{-5}$) under our default training recipe, its coarse-tail residualization provides a substantially wider stability margin in moderately aggressive regimes. Scaling beyond this threshold may require coordinating higher learning rates with complementary stabilization techniques (such as larger batch sizes or tighter clipping thresholds), which we leave to future work.
\vspace{-4mm}

\subsection{Comparing the two-control endpoints}
\label{sec:control-comparison}

The analysis in Sec.~\ref{sec:unified} identifies a distinct opportunity to
reduce within-tail variance through $\gamma$. We evaluate that prediction
using SC-TA ($\gamma=1$) and ResOPD ($\gamma=0$), keeping the observable
coarse control $\bm B=\bmean_B$ fixed.

\subsubsection{Cross-distribution fixed-prefix audit}
While the two-control framework indicates that SC-TA ($\gamma=1$) can mitigate within-tail fluctuations beyond ResOPD ($\gamma=0$), Equation~(\ref{eq:event-improvement}) implies this ordering is distribution-dependent rather than universal. To evaluate stability across domains, we freeze the Qwen3.5-4B/27B pair (Top-4 support, 64-token horizon) across three natural prompt distributions (DeepMath \citep{deepmath}, MegaScience \citep{fan2025megascience} and SynLogic \citep{liu2026synlogic}), evaluating 128 trajectories (8,192 prefixes) with 20,000 bootstrap resamples. We track the normalized trace variance reduction:
\begin{equation}
  \Delta \coloneqq \frac{\mathcal{V}_{\mathrm{SC}} - \mathcal{V}_{\mathrm{ResOPD}}}{\mathcal{V}_{\mathrm{sampled}}},
\end{equation}
where negative values favor SC-TA.

As reported in Tab.~\ref{tab:scta-cross-distribution}, the aggregate trace variance consistently favors SC-TA over ResOPD across all three evaluated domains ($\Delta \in [-0.207, -0.126]$). Simultaneously, prefix-level crossover remains non-trivial (up to $18.0\%$ of prefixes favor ResOPD in SynLogic). This empirically corroborates the qualified prediction of our two-control formulation: while setting $\gamma=1$ may yield lower variance in expectation, the local variance ordering remains strictly prefix-dependent rather than globally uniform.

In principle, one could pre-select a subset of prompts from the training distribution to empirically calibrate an optimal coefficient $\gamma^\star$, potentially unlocking additional variance reduction during optimization (even though $\gamma^\star$ inherently evolves alongside model weights). In practice, however, our experiments indicate that the incremental gains from such calibration over the fixed heuristic $\gamma=1$ are marginal. We therefore omit further discussion in favor of simplicity.

\begin{table}[t]
\caption{\textbf{SC-TA vs.\ ResOPD covariance trace ordering across natural prompt distributions.}}
\vspace{-2mm}
\label{tab:scta-cross-distribution}
\centering
\small
\setlength{\tabcolsep}{6.0pt}
\begin{tabular}{@{}lccc@{}}
\toprule
Dataset & $\mathcal{V}_{\mathrm{SC}} / \mathcal{V}_{\mathrm{ResOPD}}$ & $\Delta$ & ResOPD-Lower Prefixes \\
\midrule
DeepMath    & 0.512 & $-0.126$ & \phantom{0}6.8\% \\
MegaScience & 0.532 & $-0.207$ & 11.3\% \\
SynLogic    & 0.558 & $-0.161$ & 18.0\% \\
\bottomrule
\end{tabular}
\vspace{-2mm}
\end{table}

\begin{table}[t]
\caption{\textbf{Matched evaluation curves across training updates (\%).}
Evaluated on Qwen3.5-27B-to-4B distillation across AIME26/HMMT26 (avg@16) and held-out DeepMath/OlymMATH (avg@2).}
\vspace{-2mm}
\label{tab:scta-curve}
\centering
\small
\setlength{\tabcolsep}{5.5pt}
\begin{tabular}{@{}lcccccc@{}}
\toprule
Method & Base (Step 0) & Step 6 & Step 32 & Step 64 & Step 128 & Step 192 \\
\midrule
HETS  & 42.85 & \textbf{48.86} & 45.04 & 45.94 & 46.60 & 45.72 \\
ResOPD   & 42.85 & 48.23 & \textbf{47.95} & 47.10 & 47.71 & \textbf{50.12} \\
SC-TA & 42.85 & 47.66 & 46.69 & \textbf{47.26} & \textbf{48.27} & 46.33 \\
\bottomrule
\end{tabular}
\vspace{-4mm}
\end{table}

\subsubsection{Downstream evaluation comparison}
While SC-TA halves the fixed-prefix residual covariance trace of ResOPD, this advantage does not necessarily translate into superior downstream accuracy. As shown in Tab.~\ref{tab:scta-curve},
SC-TA does not exhibit a consistent downstream advantage over ResOPD: it is slightly stronger at several intermediate checkpoints, whereas ResOPD achieves the higher terminal score, with trajectory-level AUC remaining similar.

This decoupling between gradient covariance trace and downstream performance is informative rather than contradictory. While lower estimator variance validates the two-control residualization mechanism locally, downstream benchmark accuracy reflects the global trajectory of policy updates.
Consequently, while SC-TA exhibits lower overall variance across the evaluated audit sets, ResOPD retains strong aggregate performance in this evaluated setting without explicitly subtracting the additional within-tail score fluctuation.
\vspace{-4mm}

\FloatBarrier
\section{Conclusion}

We introduced ResOPD, a principled estimator for full-vocabulary reverse-KL distillation under sparse teacher feedback. By integrating the observable coarse distribution analytically and sampling only the unresolved fine-tail residual, ResOPD reduces variance in the evaluated settings while preserving exact unbiasedness under ancestral on-policy sampling, without expanding the teacher payload or adding forward passes. Supported by theoretical analysis, exact covariance audits, and empirical evaluations across math, science, and coding tasks, ResOPD demonstrates higher aggregate downstream scores in the evaluated settings.


\FloatBarrier

\Needspace{5\baselineskip}
\renewcommand{\bibfont}{\small}
\setlength{\bibsep}{3pt plus 1pt minus 1pt}
\bibliographystyle{plainnat}
\bibliography{refs}

\clearpage
\appendix
\section{Detailed estimator derivations}
\label{app:theory}

The main text gives the estimator, its unbiasedness identity, and the unified
variance analysis. This appendix supplies the detailed Top-$k$ bias
derivations and the single-control oracle calculation, including the
boundary-term cancellation for the complete coarse distribution.

\subsection{Three biased forms of Teacher \texorpdfstring{Top-$k$}{Top-k} reverse KL}
\label{app:topk-bias}

This subsection distinguishes three operations that are often conflated:
renormalizing both distributions on the head, truncating the scalar reverse-KL objective, and truncating only the stopped-coefficient gradient. Each is
computable from Teacher Top-$k$, but none generally estimates correctly the gradient of full-vocabulary reverse KL. Here ``bias'' is relative to that target gradient, not Monte Carlo sampling bias. Fix the support $S$ and Teacher probabilities,
let $z$ be unrestricted Student logits, and define
\begin{equation}
 \begin{gathered}
 R\coloneqq\sum_i p_i r_i=\KL(p\|q),\qquad
 R_S\coloneqq\sum_{i\in S}p_i r_i,\\
 P\coloneqq\sum_{i\in S}p_i,\qquad
 Q\coloneqq\sum_{i\in S}q_i.
 \end{gathered}
\end{equation}
The exact full-rKL logit gradient, which provides the reference below, is
\begin{equation}
 [\nabla_z R]_j=p_j(r_j-R).
 \label{eq:full-rkl-logit-gradient}
\end{equation}

\paragraph{Renormalized Top-$k$ reverse KL.}
Define $\widetilde p_i=p_i/P$, $\widetilde q_i=q_i/Q$ and
$\widetilde r_i=\log(\widetilde p_i/\widetilde q_i)$ for $i\in S$, with
$\widetilde R=\sum_{i\in S}\widetilde p_i\widetilde r_i$. Since
$\widetilde p_i=\exp (z_i)/\sum_{\ell\in S}\exp (z_\ell)$, the objective and its
logit gradient are
\begin{equation}
 \begin{aligned}
 L_{\mathrm{renormalized}}
 &=\sum_{i\in S}\widetilde p_i
   \log\frac{\widetilde p_i}{\widetilde q_i},\\
 [\nabla_z L_{\mathrm{renormalized}}]_j
 &=\ind\{j\in S\}\widetilde p_j
   (\widetilde r_j-\widetilde R).
 \end{aligned}
 \label{eq:renorm-topk-gradient}
\end{equation}
Every tail-logit component is exactly zero, whereas
Equation~(\ref{eq:full-rkl-logit-gradient}) is generally nonzero there. The
global minimum is the set
\begin{equation}
 \widetilde p_i=\widetilde q_i
 \quad(i\in S)
 \quad\Longleftrightarrow\quad
 p_i=\frac{P}{Q}q_i\quad(i\in S),
 \label{eq:renorm-topk-optimum}
\end{equation}
with arbitrary $P$ and arbitrary tail shape. Renormalization therefore learns
only the conditional head distribution; it identifies neither head mass nor
the omitted conditional distribution.

\paragraph{Scalar-truncated Top-$k$ reverse KL.}
The unnormalized scalar shortcut is
$L_{\mathrm{truncated}}=R_S=\sum_{i\in S}p_i r_i$. Writing
$C_S\coloneqq\sum_{i\in S}p_i(r_i+1)=R_S+P$, direct differentiation gives
\begin{equation}
 \begin{aligned}
 \nabla_\theta L_{\mathrm{truncated}}
 &=\bhead+\nabla_\theta P,\\
 [\nabla_z L_{\mathrm{truncated}}]_j
 &=\ind\{j\in S\}p_j(r_j+1)-p_jC_S.
 \end{aligned}
 \label{eq:scalar-truncated-gradient}
\end{equation}
Thus its bias relative to full rKL is
$\nabla_\theta P-\tau_p\bmean_{\mathrm{tail}}$. In particular, for
$j\notin S$, $[\nabla_z L_{\mathrm{tr}}]_j=-p_jC_S$; when $C_S>0$,
gradient descent increases every omitted logit.

The corresponding global optimum makes the failure more explicit. For each
head coordinate,
\begin{equation}
 x\log\frac{x}{q_i}\geq-\frac{q_i}{e},
 \qquad\text{with equality iff }x=\frac{q_i}{e}.
\end{equation}
Because $\sum_{i\in S}q_i/e=Q/e<1$, all coordinatewise minima are jointly
feasible, and therefore
\begin{equation}
 p_i^\star=\frac{q_i}{e}\quad(i\in S),\qquad
 P^\star=\frac{Q}{e},\qquad
 \sum_{j\notin S}p_j^\star=1-\frac{Q}{e},
 \label{eq:scalar-truncated-optimum}
\end{equation}
with arbitrary allocation inside the tail. Even if $Q$ is near one, this
objective prefers roughly $1-1/e$ Student mass outside the Teacher head; the
tail-push is thus an objective-level property, not merely a transient gradient
sign.

\paragraph{Stopped-coefficient Top-$k$ reverse-KL gradient.}
Stopping the coefficients removes the $\nabla P$ boundary term, yet leaves the tail expectation uncorrected:
\begin{equation}
  \begin{gathered}
    \bgrad_{\text{Top-}k} \coloneqq \sum_{i \in S} \operatorname{stopgrad}(p_i r_i) \bscore_i = \bhead, \\
    [\bgrad_{\text{Top-}k}]_j = \ind\{j \in S\} p_j r_j - p_j R_S, \qquad
    \bgrad_{\text{Top-}k} - \bgrad = -\tau_p \bmean_{\mathrm{tail}}.
  \end{gathered}
  \label{eq:stopped-topk-gradient-bias}
\end{equation}
In unrestricted categorical-logit space with a non-empty tail, vanishing gradients ($\bgrad_{\text{Top-}k} = \mathbf{0}$) force $R_S = 0$ via the tail components ($j \notin S$), which subsequently implies $r_i = 0$ for all head components ($i \in S$). Consequently, $p_i = q_i$ on the head, correctly matching the aggregate tail mass ($1 - P = 1 - Q$); however, the conditional distribution over the tail remains unconstrained and arbitrary. While this stopped-gradient formulation alters the underlying target the least among the three heuristics, it remains inherently biased whenever the tail vector is non-zero. HETS and ResOPD resolve this residual bias by introducing a sampled tail correction to restore the missing expectation.

\subsection{Single-control residualization and its variance optimum}
\label{app:single-control}

Use the head/tail notation of Equation~(\ref{eq:head-tail-definitions}), let
$J=\ind\{A\notin S\}$, and let $\bm B$ be any finite prefix-measurable vector
stopped before sampling $A\sim p$. Consider the one-control estimator
\begin{equation}
 \widehat{\bgrad}(\bm B)
 \coloneqq
 \bhead+\tau_p\bm B+J(\btoken_A-\bm B).
 \label{eq:single-control-estimator}
\end{equation}
It is unbiased for every such $\bm B$:
\begin{equation}
 \E[\widehat{\bgrad}(\bm B)]
 =\bhead+\tau_p\bm B
  +\tau_p(\bmean_{\mathrm{tail}}-\bm B)
 =\bgrad.
 \label{eq:single-control-unbiased}
\end{equation}
Conditioned on $J=0$, its value is the deterministic vector
$\bhead+\tau_p\bm B$. Conditioned on $J=1$, its mean is
$\bhead+\tau_p\bm B+\bmean_{\mathrm{tail}}-\bm B$ and its covariance is
$\Cov_{\mathrm{tail}}(\btoken_A)$. The two conditional means differ by
$\bmean_{\mathrm{tail}}-\bm B$. Hence the law of total covariance gives
\begin{equation}
 \mathcal V(\widehat{\bgrad}(\bm B))
 =\tau_p\mathcal V_{\mathrm{tail}}(\btoken_A)
 +P\tau_p\lVert\bmean_{\mathrm{tail}}-\bm B\rVert_2^2.
 \label{eq:single-control-variance}
\end{equation}
The first term is independent of $\bm B$. Therefore, whenever
$0<P<1$, the unique trace-covariance minimizer and its exact regret are
\begin{equation}
 \begin{aligned}
 \bm B^\star&=\bmean_{\mathrm{tail}},\\
 \mathcal V(\widehat{\bgrad}(\bm B))
 -\mathcal V(\widehat{\bgrad}(\bmean_{\mathrm{tail}}))
 &=P\tau_p\lVert\bm B-\bmean_{\mathrm{tail}}\rVert_2^2.
 \end{aligned}
 \label{eq:single-control-oracle}
\end{equation}
The oracle removes all head-versus-tail event variance, but it is unavailable
under a sparse Teacher interface because it requires the full conditional
fine-tail mean.

HETS and ResOPD are the two payload-compatible choices
\begin{equation}
 \widehat{\bgrad}^{\mathrm{HETS}}=\widehat{\bgrad}(\bm 0),
 \qquad
 \widehat{\bgrad}^{\mathrm{ResOPD}}=\widehat{\bgrad}(\bmean_B).
\end{equation}
Their regrets relative to the single-control oracle are respectively
$P\tau_p\lVert\bmean_{\mathrm{tail}}\rVert_2^2$ and
$P\tau_p\lVert\bmean_{\mathrm{tail}}-\bmean_B\rVert_2^2$. Subtraction yields
\begin{equation}
  \mathcal{V}(\widehat{\bgrad}^{\mathrm{HETS}}) - \mathcal{V}(\widehat{\bgrad}^{\mathrm{ResOPD}}) = P\tau_p \!\left( 2\langle\bmean_{\mathrm{tail}}, \bmean_B\rangle - \lVert\bmean_B\rVert_2^2 \right).
  \label{eq:hets-cfr-variance-appendix}
\end{equation}
Thus ResOPD has lower trace covariance exactly when $\bmean_B$ is closer to
$\bmean_{\mathrm{tail}}$ than the HETS control $\bm 0$ is; no unconditional
ordering is asserted.

It remains to verify why $\bmean_B$ is available without the fine Teacher
tail. The reverse KL of the complete coarse distribution over the $k$ head
tokens and one tail bucket is
\begin{equation}
 L_{\mathrm{coarse}}
 \coloneqq
 \sum_{i\in S}p_i\log\frac{p_i}{q_i}
 +\tau_p\log\frac{\tau_p}{\tau_q}.
\end{equation}
Since
$\nabla\tau_p=\tau_p\bar{\bscore}_{\mathrm{tail}}=-\nabla P$,
direct differentiation
gives
\begin{equation}
 \begin{aligned}
 \nabla L_{\mathrm{coarse}}
 &=\bhead+\nabla P+(b+1)\nabla\tau_p\\
 &=\bhead+b\nabla\tau_p
  =\bhead+\tau_p\bmean_B
  =\bgrad_{\mathrm{coarse}}.
 \end{aligned}
\end{equation}
The boundary terms cancel because the tail pseudo-token completes the coarse
support. This is why $\bgrad_{\mathrm{coarse}}$ is an exact gradient (but biased), even though
$\bmean_B$ need not equal the true conditional fine-tail mean.

\FloatBarrier
\section{Detailed training and evaluation configurations}
\label{app:hyperparameters}

Tabs.~\ref{tab:opd-hyperparameters} and~\ref{tab:qwen3-opd-hyperparameters}
collect the complete settings for the primary mathematics and cross-domain
runs. Their sampling assumptions are discussed in
Secs.~\ref{subsec:evidence} and~\ref{sec:ex-setup}; the matched
full-support mathematics ablation is reported in
Sec.~\ref{sec:rollout-ablation}.

\begin{table}[t]
\caption{\textbf{Primary Math OPD experimental configurations and hyperparameters.}}
\label{tab:opd-hyperparameters}
\centering
\small
\renewcommand{\arraystretch}{1.18}
\setlength{\tabcolsep}{6pt}
\begin{tabular}{@{}lll@{}}
\toprule
\textbf{Category} & \textbf{Hyperparameter} & \textbf{Configuration / Value} \\
\midrule

\multirow{2}{*}{\textbf{Models}}
& Student model & Qwen3.5-4B \\
& Teacher model & Qwen3.5-27B \\
\midrule

\multirow{2}{*}{\textbf{Data}}
& Training dataset & 1{,}024 prompts sampled from DeepMath-103K \\
& Max prompt length & 2{,}048 \\
\midrule

\multirow{5}{*}{\textbf{Sampling}}
& Enable thinking & True \\
& Max response length & 32{,}768 (training and evaluation) \\
& temperature & 1.0 \\
& top-$p$ & 0.95 \\
& top-$k$ & 20 \\
\midrule

\multirow{5}{*}{\textbf{Training}}
& Training epochs & 3 \\
& Global batch size & 16 \\
& Optimizer & AdamW \\
& Gradient clipping & 1.0 \\
& Learning rate & $1\times10^{-6}$ \\
\midrule

\multirow{4}{*}{\textbf{System \& Hardware}}
& Compute infrastructure & $16\times$ NVIDIA H200 GPUs \\
& Resource allocation & 8 GPUs for Student rollout, 4 for Student training, \\
& & 4 for hosting the Teacher inference service \\
& Software frameworks & Training: verl \quad Inference: vLLM \\
\bottomrule
\end{tabular}
\end{table}


\section{Step-wise evaluation results}

\subsection{Math OPD experiments}
\label{app:math-learning-trajectories}

\begin{table}[t]
\caption{\textbf{Cross-domain Qwen3 OPD experimental configurations and hyperparameters.}}
\label{tab:qwen3-opd-hyperparameters}
\centering
\small
\renewcommand{\arraystretch}{1.18}
\setlength{\tabcolsep}{6pt}
\begin{tabular}{@{}lll@{}}
\toprule
\textbf{Category} & \textbf{Hyperparameter} & \textbf{Configuration / Value} \\
\midrule
\multirow{2}{*}{\textbf{Models}}
& Student model & Qwen3-4B \\
& Teacher model & Qwen3-30B-A3B-Instruct-2507 \\
\midrule
\multirow{3}{*}{\textbf{Data}}
& Training dataset & 30,000 deduplicated prompts; \\
& & Math/Code/Science: 10,000 each \\
& Max prompt length & 1,024 \\
\midrule
\multirow{5}{*}{\textbf{Sampling}}
& Enable thinking & False \\
& Max response length & 8192 (training) and 32768 (evaluation) \\
& temperature & 1.0 \\
& top-$p$ & 1.0 \\
& top-$k$ & $-1$ \\
\midrule
\multirow{7}{*}{\textbf{Training}}
& Training steps & 100 \\
& Global batch size & 288 \\
& Optimizer & AdamW \\
& Gradient clipping & 1.0 \\
& Peak learning rate & $5\times10^{-6}$ \\
& LR schedule & 10-step warmup; \\
& & cosine decay to $0.1\times$ peak \\
\midrule
\multirow{4}{*}{\textbf{System \& Hardware}}
& Compute infrastructure & $16\times$ NVIDIA H200 GPUs \\
& Resource allocation & 8 GPUs for student rollout, 4 for student training, \\
& & 4 for hosting the teacher inference service \\
& Software frameworks & Training: verl \quad Inference: vLLM \\
\bottomrule
\end{tabular}
\end{table}

\begin{figure}[t]
\centering
\includegraphics[width=0.85\textwidth]{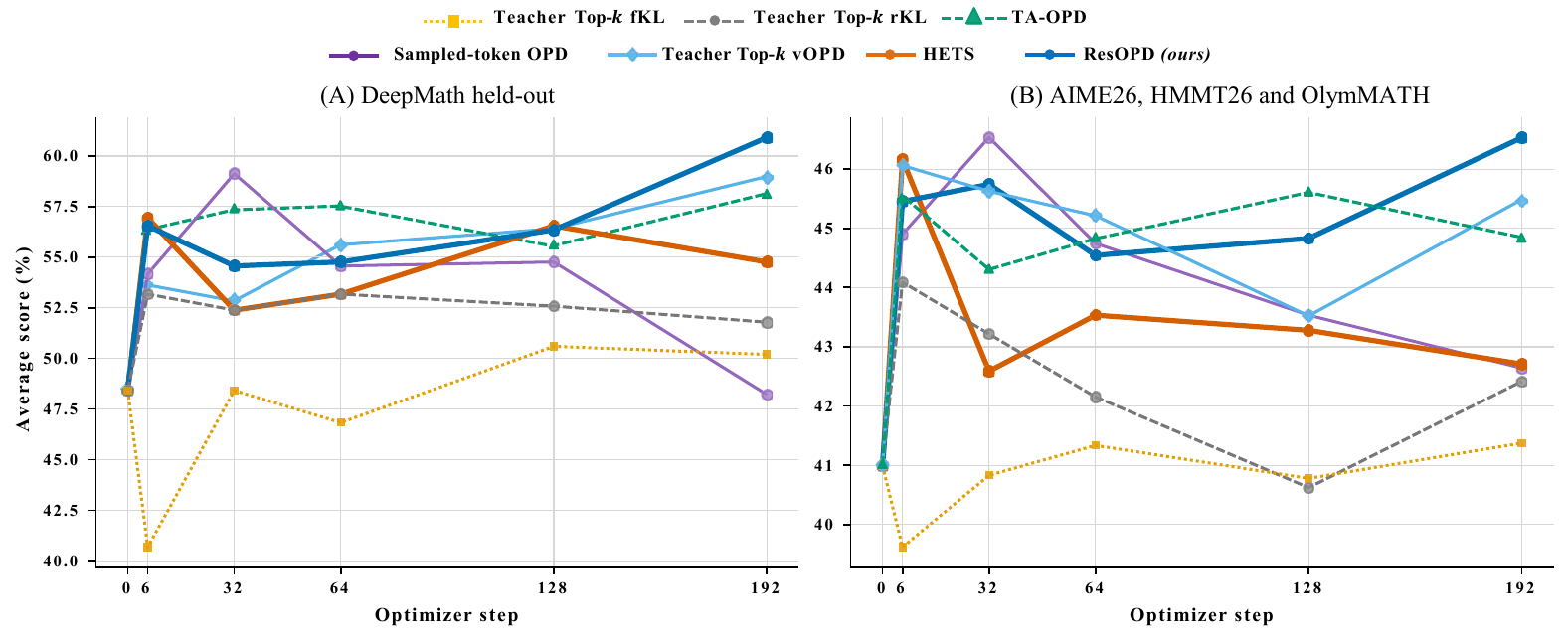}
\caption{\textbf{Step-wise learning curves} corresponding to Tab.~\ref{tab:math-terminal} on DeepMath held-out (left) and external competitions (right) across 192 optimizer steps.}
\label{fig:estimator-learning}
\end{figure}

Fig.~\ref{fig:estimator-learning} illustrates the training trajectories across evaluation suites. 
On the DeepMath held-out set, \method{} displays consistent late-stage gains, achieving the highest terminal accuracy among all estimators. 
While competitive baselines such as TA-OPD and Teacher Top-$k$ vOPD show steady improvements, they fall short of \method{} at the terminal checkpoint. Meanwhile, Sampled-token OPD deteriorates severely after an early peak, and HETS suffers a late-stage drop. 
A similar trend holds on the external math competitions: although TA-OPD maintains strong intermediate trajectory performance and vOPD remains competitive, \method{} exhibits superior late-stage resilience and decisively surpasses all alternatives at the final step, avoiding the post-peak degradation observed in estimators like Sampled-token OPD and HETS.

Together with the lower fixed-prefix covariance trace and improved online
training diagnostics reported elsewhere in the paper, these trajectories
are consistent with the proposed coarse-tail residualization reducing
stochastic gradient noise and improving late-stage optimization stability.

\subsection{Cross-Domain OPD Experiments}
\label{app:cross-qwen3}

\begin{figure}[htbp]
  \centering
  \includegraphics[width=0.85\columnwidth]{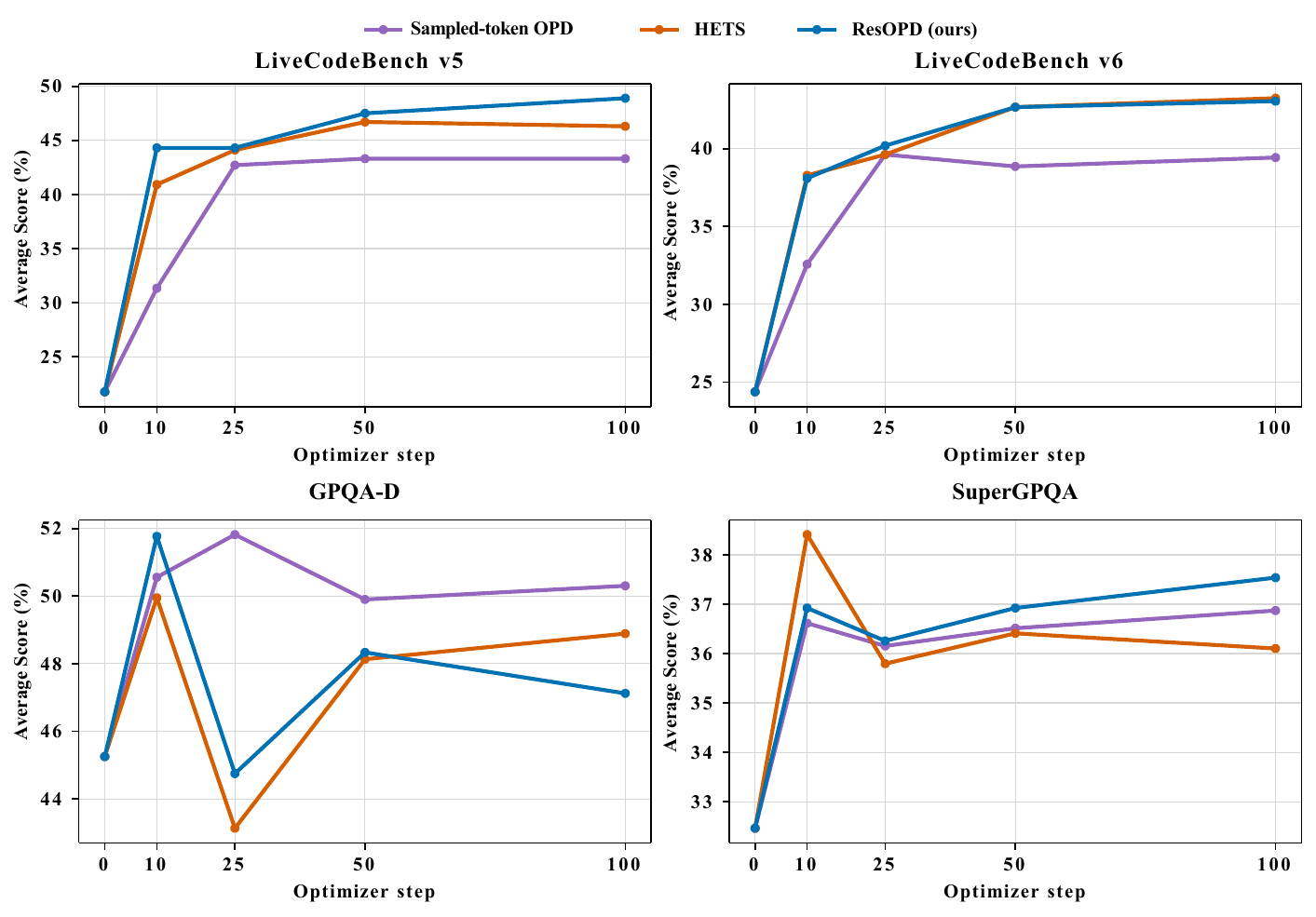}
    \caption{\textbf{Evaluation trajectories during cross-domain transfer distillation} from Qwen3-30B-A3B-Instruct to Qwen3-4B across optimizer steps (0, 10, 25, 50, and 100).}
    \label{fig:qwen3-transfer-curves}
\end{figure}

Fig.~\ref{fig:qwen3-transfer-curves} illustrates the checkpoint evaluation trajectories along the fixed training progression (steps 0, 10, 25, 50, and 100) across all four benchmarks. On code benchmarks (LiveCodeBench v5 and v6), ResOPD and HETS rapidly converge and consistently outperform Sampled-token OPD by a substantial margin throughout training, with ResOPD achieving the highest overall trajectory AUC. In contrast, science benchmarks display mixed domain dynamics: Sampled-token OPD demonstrates superior retention and final accuracy on GPQA-D, whereas ResOPD remains competitive across training steps and achieves the best terminal performance on SuperGPQA. These trajectories corroborate that while Sampled-token OPD remains strong in specialized scientific reasoning, ResOPD provides transfer performance on code tasks without sacrificing overall multi-domain balance.

\FloatBarrier
\section{Details of diagnostics}
\label{app:math-curves}

\subsection{What the online residual moments measure}
\label{app:online-residual-moments}

At optimizer update $u$, let $i \in \{1, \dots, B_u\}$ index the non-empty sampled responses and $t \in \{1, \dots, T_{ui}\}$ index their valid tokens. We define the scalar support residual and its components as
\begin{equation}
  P_{uit} \coloneqq p_\theta(S_{uit} \mid x_{uit}), \qquad
  H_{uit} \coloneqq \ind\{A_{uit} \in S_{uit}\}, \qquad
  Z_{uit} \coloneqq H_{uit} - P_{uit}.
  \label{eq:online-support-residual}
\end{equation}
Here, $P_{uit}$ denotes the Student probability mass allocated to the transmitted Teacher \text{Top-}$k$ support $S_{uit}$, while $H_{uit}$ indicates whether the sampled Student token lands within this set. Under an exact on-policy draw, the residual is conditionally zero-mean, $\E[Z_{uit} \mid x_{uit}] = 0$, with conditional variance $\E[Z_{uit}^2 \mid x_{uit}] = P_{uit}(1 - P_{uit})$. More generally, its uncentered second moment captures both intrinsic sampling variability and any realized calibration drift.

Fig.~\ref{fig:training-dynamics-event-g0} reports two diagnostic aggregations of this quantity. Panel~C displays the masked token-level second moment:
\begin{equation}
  M_{2,\mathrm{tok}}^{(u)} = \frac{1}{\sum_{i=1}^{B_u} T_{ui}} \sum_{i=1}^{B_u} \sum_{t=1}^{T_{ui}} Z_{uit}^2,
  \label{eq:online-token-m2}
\end{equation}
whereas Panel~D evaluates the trajectory-level variance by first averaging within each response and then weighting responses uniformly:
\begin{equation}
  \bar{Z}_{ui} = \frac{1}{T_{ui}} \sum_{t=1}^{T_{ui}} Z_{uit}, \qquad
  M_{2,\mathrm{traj}}^{(u)} = \frac{1}{B_u} \sum_{i=1}^{B_u} \bar{Z}_{ui}^2.
  \label{eq:online-trajectory-m2}
\end{equation}
Intuitively, $M_{2,\mathrm{tok}}$ captures the instantaneous magnitude of the unresolved head-versus-tail events (which is leveraged in Fig.~\ref{fig:topk-ablation} to approximate the between-event variance factor $P\tau_p$), whereas $M_{2,\mathrm{traj}}$ measures the persistent, signed event imbalance surviving along sequence rollouts. Defining the batch mean $\bar{Z}_u = B_u^{-1} \sum_{i=1}^{B_u} \bar{Z}_{ui}$, the empirical standard error of the trajectory mean is given by
\begin{equation}
  \widehat{\operatorname{SE}}(\bar{Z}_u) = \sqrt{\frac{\max\{M_{2,\mathrm{traj}}^{(u)} - \bar{Z}_u^2, 0\}}{B_u - 1}}.
  \label{eq:online-trajectory-se}
\end{equation}

The connection to ResOPD is exact at the scalar-coefficient level. Noting that tail indicators and tail masses satisfy $J_{uit} = 1 - H_{uit}$ and $\tau_{p,uit} = 1 - P_{uit}$, we obtain $Z_{uit} = \tau_{p,uit} - J_{uit}$. Substituting this into Equation~(\ref{eq:residualization}) yields
\begin{equation}
  \widehat{\bgrad}_{uit}^{\mathrm{ResOPD}} - \widehat{\bgrad}_{uit}^{\mathrm{HETS}} = Z_{uit} \bmean_{B,uit}.
  \label{eq:online-residual-cfr-link}
\end{equation}
Hence, the two $M_2$ metrics directly track the magnitude and sequence-level persistence of the random scalar modulating ResOPD's bucket control. Because they intentionally abstract away $\lVert\bmean_{B,uit}\rVert$, its alignment with the fine-tail mean, and the orthogonal within-tail fluctuations, they serve as computationally lightweight, mechanism-aligned online diagnostics rather than full covariance estimators or replacements for fixed-prefix NSR. Finally, while $Z$ can trivially be recorded for a sparse Top-$k$ objective, direct Top-$k$ rKL lacks a sampled tail correction and remains inherently biased with respect to the true rKL target. Its zero action-sampling variance is therefore not comparable to the unbiased vOPD, HETS and ResOPD estimators in Panels~C and~D.


\subsection{Finite gradient-enumeration check}
\label{app:finite-check}

We also verify the unbiasedness identities without Monte Carlo error on a
four-action softmax. The Student and Teacher probabilities are respectively
$(0.35,0.30,0.20,0.15)$ and $(0.60,0.25,0.10,0.05)$, and the first two actions form the
Teacher Top-2 support. We differentiate the full reverse KL directly, enumerate
each possible Student action, and weight its estimator gradient by the exact
Student action probability. The resulting HETS expectation matches the dense
full-rKL gradient to maximum absolute error $5.6\times10^{-17}$.

ResOPD changes the per-action vector by
$(\tau_p-J)\bmean_B$ relative to HETS. Because the same enumeration has
$\E[J]=\tau_p$, this correction sums exactly to zero and ResOPD recovers the same
dense gradient. This finite check complements the symbolic identity in
Equation~(\ref{eq:residualization}): it validates the stopped-coefficient and
support-mask implementation semantics, but it is not empirical evidence about
covariance or downstream accuracy.

\subsection{Complete frozen-prefix support diagnostics}
\label{app:fixed-prefix-sweep}

\begin{table}[t]
\caption{\textbf{Support-size ($k$) sweep under fixed Teacher Top-$k$ payloads.}
``Var.~Removed'' measures the empirical covariance-trace reduction relative to sampled-token OPD. Payload accounts for $k$ Top-$k$ entries plus the sampled token (INT32 ID and FP32 log-probability each). Student mass and Head Hit denote the predicted and empirical probabilities of landing in the Top-$k$ support, respectively. Best variance reduction and NSR per setting are in \textbf{bold}.}
\label{tab:fixed-prefix-sweep}
\centering
\small
\renewcommand{\arraystretch}{1.16}
\setlength{\tabcolsep}{4.0pt}

\begin{tabular*}{\textwidth}{@{\extracolsep{\fill}}lrrccrrrr@{}}
\toprule
\multirow{2}{*}{\shortstack[l]{Dataset \\ Student/Teacher}}
& \multirow{2}{*}{$k$}
& \multirow{2}{*}{\shortstack{Payload \\ (B/token)}}
& \multirow{2}{*}{\shortstack{Student \\ mass}}
& \multirow{2}{*}{\shortstack{Head \\ Hit}}
& \multicolumn{2}{c}{Var. Removed (\%) $\uparrow$}
& \multicolumn{2}{c}{NSR $\downarrow$} \\
\cmidrule(lr){6-7}\cmidrule(l){8-9}
& & & & & HETS & ResOPD & HETS & ResOPD \\
\midrule
\multirow{4}{*}{\shortstack[l]{\textbf{GSM8K} \\ 0.8B/4B}}
& 1  & 16  & $75.0\%$ & $74.8\%$  & $-0.5$ & $\mathbf{20.5}$ & 5.446 & $\mathbf{4.310}$ \\
& 4  & 40  & $93.1\%$ & $92.9\%$  & 27.3   & $\mathbf{52.5}$ & 3.940 & $\mathbf{2.573}$ \\
& 16 & 136 & $97.7\%$ & $97.9\%$  & 62.6   & $\mathbf{75.3}$ & 2.027 & $\mathbf{1.337}$ \\
& 64 & 520 & $99.3\%$ & $99.7\%$  & 85.1   & $\mathbf{89.5}$ & 0.807 & $\mathbf{0.570}$ \\
\addlinespace[0.25em]
\multirow{3}{*}{\shortstack[l]{\textbf{HMMT26} \\ 4B/27B}}
& 1  & 16  & $89.1\%$ & $89.3\%$   & $-1.0$ & $\mathbf{32.0}$ & 6.656 & $\mathbf{4.479}$ \\
& 4  & 40  & $99.1\%$ & $98.8\%$   & 55.8   & $\mathbf{75.3}$ & 2.914 & $\mathbf{1.630}$ \\
& 16 & 136 & $99.9\%$ & $100.0\%$  & 89.8   & $\mathbf{94.2}$ & 0.675 & $\mathbf{0.382}$ \\
\bottomrule
\end{tabular*}
\end{table}

\begin{figure}[t]
\centering
\includegraphics[width=0.96\textwidth]{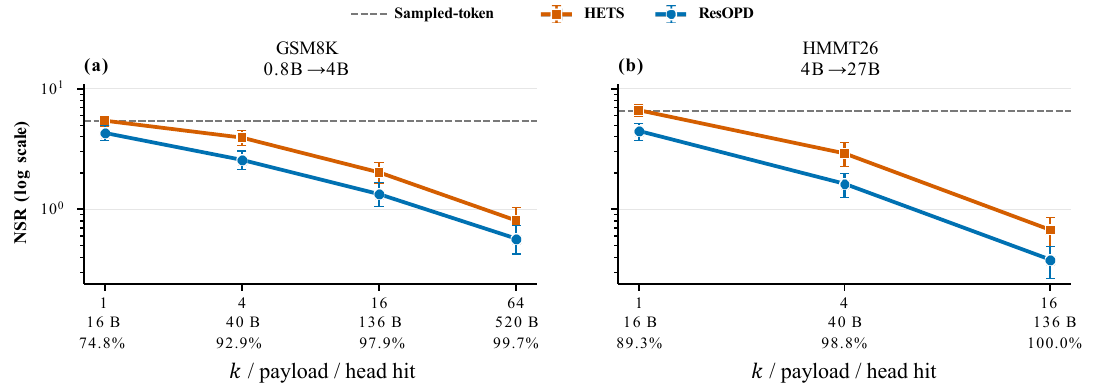}
\caption{\textbf{Comprehensive frozen-prefix support-size ($k$) diagnostics across benchmarks.} 
ResOPD consistently reduces the dimensionless NSR across all evaluated support sizes on GSM8K and HMMT (log scale). Axis annotations detail the transmitted Teacher support size $k$, the corresponding communication payload (bytes/token), and the empirical head-hit probability. Error bars denote 95\% trajectory-level bootstrap confidence intervals.}
\label{fig:fixed-prefix-diagnostics}
\end{figure}

Across both suites in Tab.~\ref{tab:fixed-prefix-sweep}, ResOPD consistently outperforms HETS at every matched support size, achieving higher covariance-trace elimination and lower NSR. Expanding $k$ from $4$ to $16$ and $64$ scales transmission costs by $3.4\times$ and $13\times$ ($40 \to 136 \to 520$\,B/token) for diminishing marginal variance gains. Crucially, the head hit rates demonstrate why high support coverage alone is insufficient: even when head mass reaches $98.8\%$ ($k=4$ on HMMT26), the rare $1.2\%$ tail draws still induce a substantial HETS--ResOPD variance gap ($55.8\%$ vs.\ $75.3\%$ removed). Moreover, at $k=1$ where HETS exhibits negative reduction ($-1.0\%$), ResOPD reliably eliminates $32.0\%$ of the gradient variance. This firmly corroborates the theoretical condition in Equation~(\ref{eq:hets-cfr-variance-appendix}), confirming that uncorrected tail draws dominate residual gradient noise despite their low nominal probability mass.

\subsection{Training dynamics across learning rates}
\label{app:lr-sensitivity}

\begin{figure}[t]
  \centering
  \includegraphics[width=0.9\columnwidth]{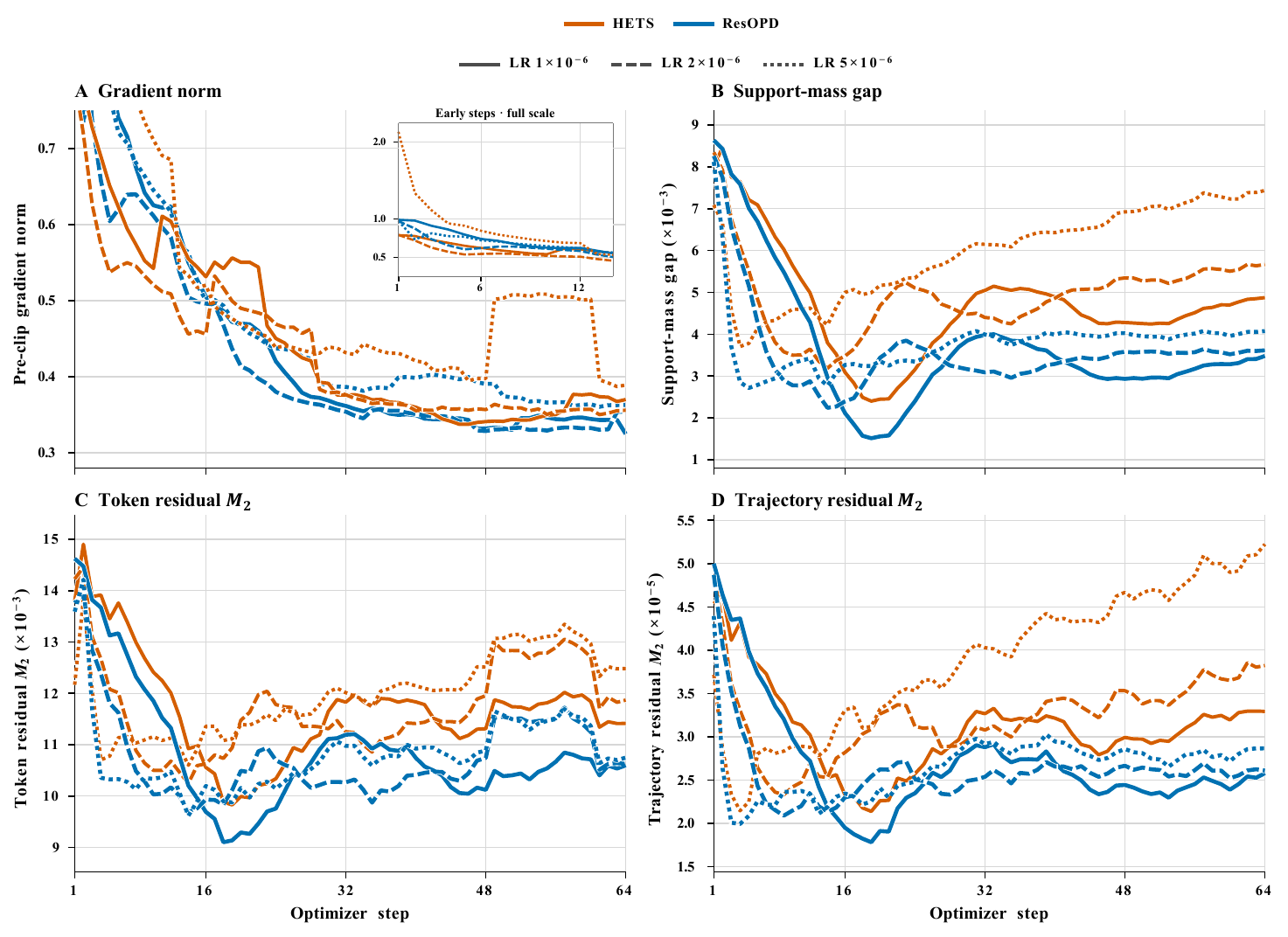}
  \caption{\textbf{Training dynamics and learning-rate sensitivity} of HETS (\textit{orange}) vs.\ ResOPD (\textit{blue}) on DeepMath across three learning rates ($1\times10^{-6}$ solid, $2\times10^{-6}$ dashed, $5\times10^{-6}$ dotted).
  \textbf{(A)} Pre-clipping global gradient norm (inset displays full-scale unclipped peaks over early steps).
  \textbf{(B)} Support-mass gap ($Q - P$) on the transmitted Teacher Top-4 tokens.
  \textbf{(C, D)} Token-level and trajectory-level second residual moments ($M_2$).
  While HETS exhibits large gradient-norm excursions and growing mass/residual discrepancies as the learning rate scales up to $5\times10^{-6}$, ResOPD maintains stable dynamics and stays close to its low-rate trajectories throughout optimization.}
  \label{fig:lr-sensitivity}
\end{figure}

\paragraph{Underlying training dynamics.}
The training trajectories in Fig.~\ref{fig:lr-sensitivity} reveal two clear mechanisms behind this divergence. First, in terms of gradient magnitude, ResOPD maintains substantially lower pre-clipping gradient norms at $5\times10^{-6}$ (mean $0.43$ vs.\ $0.48$; p99 $0.97$ vs.\ $1.84$) and completely avoids gradient clipping ($0/64$ steps vs.\ $2/64$ steps for HETS). Second, regarding residual stabilization, Panels~(B)--(D) show that ResOPD consistently curbs the support-mass gap ($Q-P$) and suppresses both token- and trajectory-level residual second moments ($M_2$), consistent with the more stable learning trajectory observed at this step size.

\end{document}